\documentclass[final,12pt]{elsarticle}

\usepackage{amssymb}
\usepackage{amsmath}
\usepackage{amsfonts}
\usepackage{booktabs}
\usepackage{multirow}
\usepackage[table]{xcolor}
\definecolor{lightblue}{HTML}{0080FF}
\usepackage{stmaryrd}
\usepackage{subcaption}
\usepackage{url}
\usepackage{pifont}
\newcommand{\cmark}{\ding{51}}%
\newcommand{\xmark}{\ding{55}}%
\usepackage{doi}
\usepackage{float}
\newcommand{\rev}[1]{{\textcolor{black}{#1}}}

\usepackage{lineno}

\journal{Automation in Construction}

\begin{document}

\begin{frontmatter}



\title{From Classification to Segmentation with Explainable AI: A Study on Crack Detection and Growth Monitoring}


\author[1]{Florent Forest}
\ead{florent.forest@epfl.ch}
\author[2]{Hugo Porta}
\ead{hugo.porta@epfl.ch}
\author[2]{Devis Tuia}
\ead{devis.tuia@epfl.ch}
\ead[url]{https://eceo.epfl.ch}
\author[1]{Olga Fink}
\ead{olga.fink@epfl.ch}
\ead[url]{https://imos.epfl.ch}
\affiliation[1]{organization={Intelligent Maintenance and Operations Systems (IMOS), EPFL},
    city={Lausanne},
    postcode={1015}, 
    country={Switzerland}}
\affiliation[2]{organization={Environmental Computational Science and Earth Observation (ECEO), EPFL},
    city={Sion},
    country={Switzerland}}

\begin{abstract}
Monitoring surface cracks in infrastructure is crucial for structural health monitoring. Automatic visual inspection offers an effective solution, especially in hard-to-reach areas. Machine learning approaches have proven their effectiveness but typically require large annotated datasets for supervised training. Once a crack is detected, monitoring its severity often demands precise segmentation of the damage. However, pixel-level annotation of images for segmentation is labor-intensive. To mitigate this cost, one can leverage explainable artificial intelligence (XAI) to derive segmentations from the explanations of a classifier, requiring only weak image-level supervision. This paper proposes applying this methodology to segment and monitor surface cracks. We evaluate the performance of various XAI methods and examine how this approach facilitates severity quantification and growth monitoring. Results reveal that while the resulting segmentation masks may exhibit lower quality than those produced by supervised methods, they remain meaningful and enable severity monitoring, thus reducing substantial labeling costs.
\end{abstract}




\end{frontmatter}


\section{Introduction}\label{sec:introduction}

Cracks may appear in various types of structures, including walls \cite{cha_deep_2017,ozgenel_performance_2018,pantoja-rosero_topo-loss_2022}, road surfaces \cite{zhang_road_2016,bang_encoderdecoder_2019}, bridges \cite{abdel-qader_analysis_2003,zaurin_integration_2009,prasanna_automated_2016,xu_automatic_2019}, tunnels \cite{nguyen_development_2018}, dams, beams \cite{hoang_detection_2018}, pipes \cite{sinha_neuro-fuzzy_2006,wu_classification_2015}, railway sleepers \cite{wang_railway_2021,rombach_contrastive_2022} and slabs \cite{wang_automated_2021}, made from different materials such as masonry, concrete, brick, stone and wood. The detection and monitoring of these cracks are essential for ensuring structural safety and play a significant role in structural health monitoring. Traditional methods to detect surface cracks involved manual visual inspections on a regular basis. However, these inspections have several drawbacks, including limited availability of human resources, service interruption (e.g., closures of railway sections or bridges), inspector subjectivity, high costs, and challenges in accessing hazardous or contaminated areas. Automatic visual inspection provides a solution to overcome these issues by enabling efficient, cost-effective and safe structural health condition monitoring of surface cracks \cite{koch_review_2015,spencer_advances_2019}. This can be achieved through the use of Unmanned Aerial Vehicles (UAVs), robots or other vehicles equipped with imaging or video capturing capabilities \cite{jiang_deep_2020,mei_cost_2020,kang_autonomous_2018}. These technologies, in conjunction with data-driven approaches based on computer vision and machine learning, enable the automatic processing of large volumes of data, thereby making the assessment more objective. Image-based approaches can, naturally, only detect surface cracks. Cracks present deep inside the material, whose detection requires other types of sensing such as ultrasound \cite{tatarinov_assessment_2019} or X-ray \cite{shayan_microscopic_1992}, are not considered in this work.

The automated detection and segmentation of cracks in images pose challenges due to the diverse aspects of cracks, the complexity and diversity of material textures, and irregular illumination conditions. Various approaches have been proposed for crack detection, including classification approaches based on supervised machine learning \cite{carvalhido_automatic_2012,prasanna_automated_2016} and deep learning, notably based on convolutional neural networks (CNNs) \cite{zhang_road_2016,cha_deep_2017,ozgenel_performance_2018,xu_automatic_2019,kang_hybrid_2020}.

A significant concern in structural health monitoring is the development and propagation of cracks over time, which can lead to increased stress and eventual failure of the structure. Once a damage has been detected, it becomes crucial to monitor the evolution of its severity to trigger timely maintenance actions and prevent catastrophic consequences. The severity of a surface crack can be quantified by measuring its width, length or area \cite{carrasco_image-based_2021,wang_automated_2021,ha_assessing_2022,yu_cracklab_2022}. These measurements can be derived from the binary segmentation mask of the crack through crack profiling techniques such as skeletonization, thinning, tracking, labeling, and width measurement \cite{yang_automatic_2018}. However, achieving a precise segmentation of the damage is necessary for accurate measurements. Several approaches have been developed for segmentation, primarily based on image processing techniques \cite{tsai_critical_2010}. These methods typically involve two steps: (1) image enhancement, including noise reduction, filtering, shading correction, etc. and (2) image binarization (thresholding) to obtain the crack segmentation \cite{lee_technique_2007,hoang_detection_2018,nguyen_development_2018,fan_road_2019,kang_hybrid_2020,han_advanced_2021,carrasco_image-based_2021}. Other approaches use Fourier or wavelet transforms, edge detection \cite{abdel-qader_analysis_2003} or the percolation method \cite{yamaguchi_image-based_2008}. Crack depth is another severity metric estimated by \cite{amarasiri_modeling_2010} using optical reflection properties. Nevertheless, image processing methods usually involve complex pipelines with multiple steps, providing handcrafted solutions for specific use cases. Moreover, these methods often struggle to handle complex cases like low contrast, distracting elements, diverse material textures and complex backgrounds.

Data-driven approaches based on supervised deep learning have demonstrated excellent performance in pixel-level semantic segmentation of cracks. Numerous approaches have leveraged CNNs \cite{yang_automatic_2018,dung_autonomous_2019,bang_encoderdecoder_2019,mei_densely_2020}, among which the popular architectures U-Net \cite{liu_computer_2019,jiang_deep_2020,lau_automated_2020,augustauskas_improved_2020,ha_assessing_2022} and DeepLabv3+ \cite{yu_cracklab_2022,sun_dma-net_2022}. In \cite{huang_surface_2018}, the authors propose a fusion of saliency cues with the image within a U-Net. However, these approaches require extensive pixel-level annotations of a large number of images, which is a labor-intensive and tedious process. As an alternative approach for severity estimation, \cite{wang_automated_2021} proposed training a classifier to directly classify severity levels without the need for a segmentation step. However, this method requires images to be labeled beforehand based on their severity level and provides more limited information.



An important barrier in the development of deep learning-based automated crack segmentation systems is the high cost associated with pixel-level annotation of large sets of images. To circumvent this challenge, \emph{unsupervised} (requiring no labels) and \emph{weakly-supervised} (requiring only image-level labels) segmentation methods have received growing attention \cite{zhang_survey_2020}. As an example of an unsupervised approach, Chow et al.~\cite{chow_anomaly_2020} tackled crack segmentation through anomaly detection using an autoencoder. However, the lack of any supervisory signal hinders the performance of such approaches, as our experiments will also demonstrate. \rev{Another alternative is transfer learning, i.e. training a supervised segmentation model on publicly available datasets, and applying it to the target task. However, this requires the availability of a representative and similar enough dataset. For crack segmentation on standard materials like pavement, concrete, stone or masonry, such datasets are available \cite{kulkarni_crackseg9k_2022}. However, there are scenarios where transfer learning is challenging to apply. First, structures can experience various damage types other than cracks, such as spalling, scaling, excretions, efflorescence and corrosion as described in \cite{huthwohl_multi-classifier_2019}, cracks and spallings in railway sleepers \cite{rombach_contrastive_2022}, steel corrosion and delamination \cite{cha_autonomous_2018}, exposed rebar, alcalo-granulate reactions, water infiltration, moisture marks \cite{dawood_computer_2018}, etc. For all such types of damages, open-source segmentation datasets are not always available, as confirmed in the survey by \cite{bianchi_visual_2022}. 
Secondly, in many practical cases, the target data comes from a different domains than the available source datasets, preventing effective transfer learning. For example, when dealing with crack aspects, sizes or material textures that significantly differ from the training data, the model will not perform well. For example, \cite{tomaszkiewicz_pre-failure_2023} observed that directly transferring a crack detection model trained on a different dataset to their images of narrow cracks was a challenge due to numerous false positives. 
These scenarios highlight the need for alternative solutions. Therefore, we explore an alternative in this work. In this paper, we focus on cracks in particular due to the availability of data and ground truth labels that enable us to evaluate the proposed methodology and severity metrics. However, the proposed methodology is applicable to any damage type.}

In our work, we focus on weakly-supervised approaches that leverage explainable artificial intelligence (XAI). Explainable AI aims at enhancing the transparency and trustworthiness of AI systems, which is crucial for safety-critical applications \cite{arrieta_explainable_2019}. Numerous methods have been developed to explain the decisions made by machine learning-based systems, particularly for deep neural network models and image data. These methods differ in the types of explanation and in the techniques utilized to produce these explanations. \emph{Feature attribution explanation} methods are indirect ways of explaining a model by calculating an importance score for each variable or feature handled by the model (such as pixels in an image) to predict the target outcome, resulting in so-called \emph{attribution maps}. The main idea is to train a classifier to classify the damages, extract \emph{explanations} of the classifier's decisions in the form of per-pixel attribution maps and derive segmentation masks from these maps. In other words, the principle is to \emph{approximate segmentation masks by explanations}. The advantage of this approach is that while annotating images for segmentation is tedious, classification labels can be obtained at a fraction of the cost.

Numerous feature attribution methods exist in literature, differing in their approach to compute relevance scores, the desired properties or constraints to be satisfied, and consequently, the quality of the resulting explanations. Seibold et al.~\cite{seibold_explanations_2022} proposed to use an XAI technique called Layer-wise Relevance Propagation (LRP) \cite{bach_pixel-wise_2015} to segment damages in magnetic tiles and sewer pipe images. However, this study focused only on one of such methods, namely the LRP technique, limiting the range of applicable network architectures (for instance, LRP cannot be used in models with skip-connections). Other explanation methods were not evaluated. Furthermore, \cite{seibold_explanations_2022} solely evaluated the resulting segmentation quality in terms of F1 score, precision and recall, but the damage severity was not further assessed, which is a major requirement in many structural health monitoring applications.

In this paper, we aim to build upon this initial study, offering several key contributions. Our main contributions include proposing a comprehensive methodology and evaluating the abilities of various explainability methods in generating high-quality segmentation masks for crack detection in masonry building walls. Additionally, we assess whether this framework enables quantification and monitoring of damage severity, such as crack width measurement, to facilitate timely decision-making. Following the XAI methods taxonomy from the literature review by Arrieta et al.~\cite{arrieta_explainable_2019}, our focus is primarily on \emph{post-hoc} feature attribution methods (also known as \emph{feature relevance} methods) and \emph{architecture modification} methods suitable for convolutional neural networks and image data. 
Post-hoc explainability methods aim at explaining the decisions of a given black-box model without inherent transparency. In this work, we evaluate and compare six post-hoc techniques: Input$\times$Gradient \cite{baehrens_how_2010}, Layerwise Relevance Propagation (LRP) \cite{bach_pixel-wise_2015}, Integrated Gradients \cite{sundararajan_axiomatic_2017}, DeepLift \cite{shrikumar_learning_2019}, DeepLiftShap and GradientShap \cite{lundberg_unified_2017}. These included techniques are widely used by practitioners; they vary  in terms of the relevance computation method and exhibit different, yet reasonable, computational runtimes. 
Additionally, our study includes one recent inherently explainable architecture modification method (B-cos networks \cite{bohle_b-cos_2022}), and one recent post-hoc method that utilizes an auxiliary network to generate attribution maps (Neural Network Explainer \cite{stalder_what_2022}). For the latter, we also propose an extension specifically tailored for classification tasks in which one class represents the absence of a foreground object. This adaptation differs from its original formulation and is particularly relevant in damage classification scenarios. Importantly, all methods compared in our study generate attribution maps at the input resolution. Indeed, a high resolution is necessary to capture the thin structure of cracks. For this reason, we did not include methods producing class activation maps (CAM) at the feature level such as CAM \cite{zhou_learning_2016} and Grad-CAM \cite{selvaraju_grad-cam_2017}, as well as weakly-supervised approaches based on global pooling layers for heatmap generation \cite{hwang_self-transfer_2016,dubost_gp-unet_2017,chatterjee_weakly-supervised_2022}.


The contributions of this work are summarized as follows:

\begin{enumerate}
    \item We propose a comprehensive methodology and evaluate the performance of various explainable artificial intelligence (XAI) methods in generating high-quality segmentation masks for cracks in masonry building wall surfaces, without requiring pixel-level annotation of images. These masks are derived from the explanations of a classifier. 
    \item We extend upon the Neural Network Explainer method \cite{stalder_what_2022} to accommodate classification tasks specifically when one class represents the absence of a foreground object (i.e., damage-free image samples in the scenario of damage classification).
    \item We propose to use damage-free images as baselines in the Integrated Gradients and DeepLift-based methods, improving the quality of their explanations.
    \item We investigate the applicability of these methods in quantifying damage severity and monitoring its progression, thereby facilitating timely decision-making. To this aim, we evaluate crack severity quantification and growth monitoring abilities using severity metrics such as the number of cracks, maximum crack width and crack area.
\end{enumerate}

The remainder of this paper is structured as follows. In Section~\ref{sec:method}, we introduce the proposed methodology to produce crack segmentation masks based on classifier explanations, as well as the explainable AI techniques used throughout our work. In addition, we derive an adaptation of the Neural Network Explainer method suitable for damage classification. The following Section~\ref{sec:experiments} presents the data, crack classification models, and experimental settings. Section~\ref{sec:results} reports the results of our experiments. Finally, Section~\ref{sec:conclusion} concludes the paper and discusses the outcomes and perspectives of this research.

\section{Proposed methodology}\label{sec:method}

In this section, we introduce the general methodology for generating crack segmentation masks without pixel-level segmentation labels. This involves utilizing classifier explanations of the explainable AI methods that are evaluated in this study and performing post-processing on the resulting attribution maps. Additionally, we propose an adaptation of the Neural Network Explainer method suitable for damage classification.

\subsection{Overview of the proposed methodology}

We propose the following methodology to generate crack segmentation masks based on classifier explanations, without the need for pixel-level annotation of ground-truth segmentation masks:

\begin{enumerate}
    \item Collect and label a dataset of positive (containing a crack) and negative (damage-free, without any crack) image patches.
    \item Train a binary classifier using the labeled training images.
    \item Perform inference on unseen test images. For each positive prediction, extract attribution maps from the classifier corresponding to the positive class, using an XAI method able to extract attribution maps at input resolution.
    \item Post-process the resulting attribution maps:
    \begin{enumerate}
        \item Binarize the continuous attributions to obtain binary masks.
        \item Apply morphological operations to close gaps in the masks and remove noisy attributions.
    \end{enumerate}
    \item Compute crack severity levels based on the resulting masks.
\end{enumerate}

The main principle is to approximate segmentation masks by classifier explanations. In cases where a well-performing classifier can be obtained for distinguishing between damage-free and damaged image samples, the explanations provided by an XAI approach, especially the pixel-level attribution maps, become valuable for generating accurate segmentation masks of damages. These maps aim at highlighting the discriminative regions that contribute to the target class. In the context of crack detection, a pixel should contribute to the crack class if and only if it is part of a cracked region. Therefore, there is an expected correspondence between explanations and segmentations, as the attribution maps should accurately highlight the regions where cracks are present.

\begin{figure}
    \centering
    \begin{subfigure}[t]{0.48\textwidth}
        \vskip 0pt
        \centering
        \includegraphics[width=\textwidth]{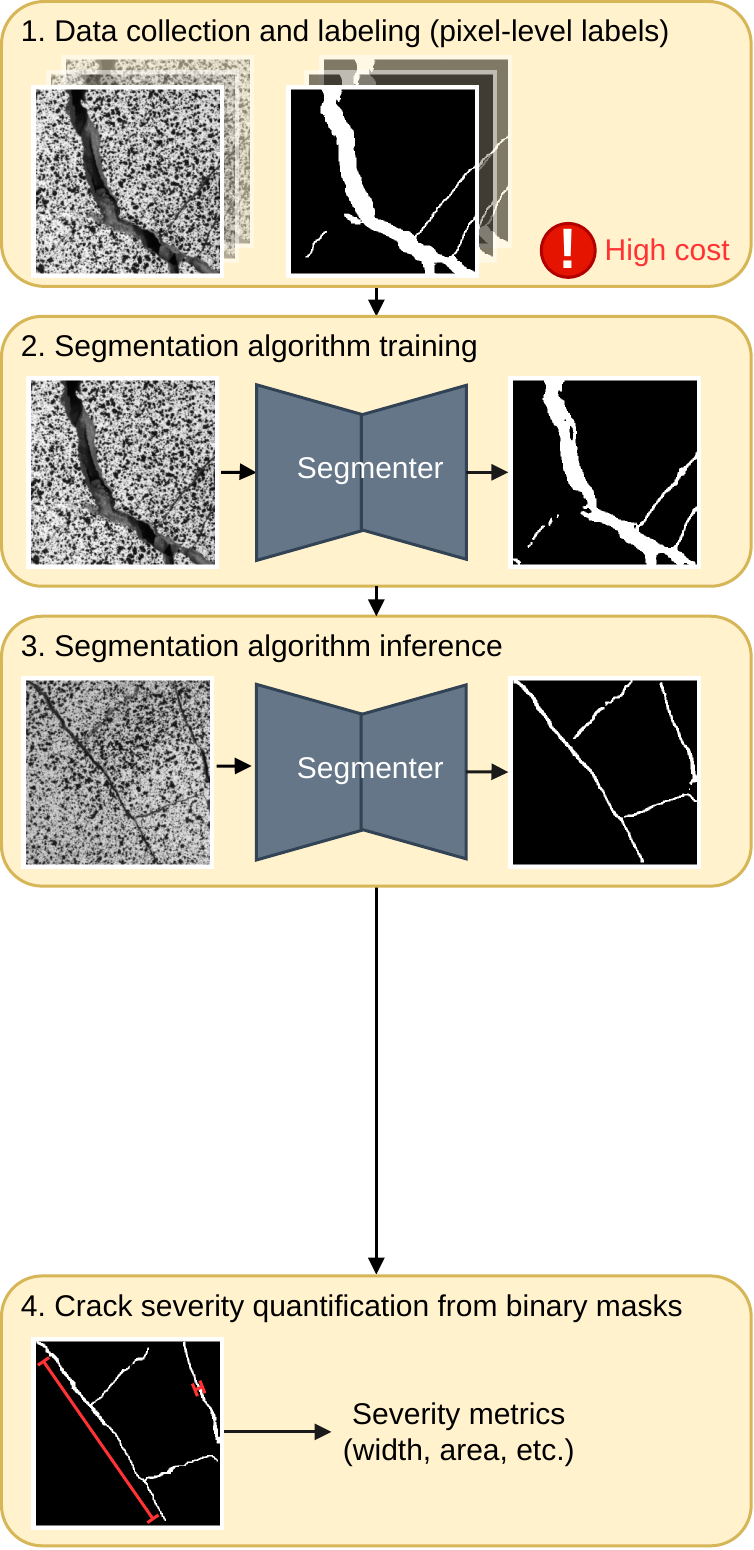}
        \caption{}
        \label{fig:workflow-seg}
    \end{subfigure}
    \begin{subfigure}[t]{0.48\textwidth}
        \vskip 0pt
        \centering
        \includegraphics[width=\textwidth]{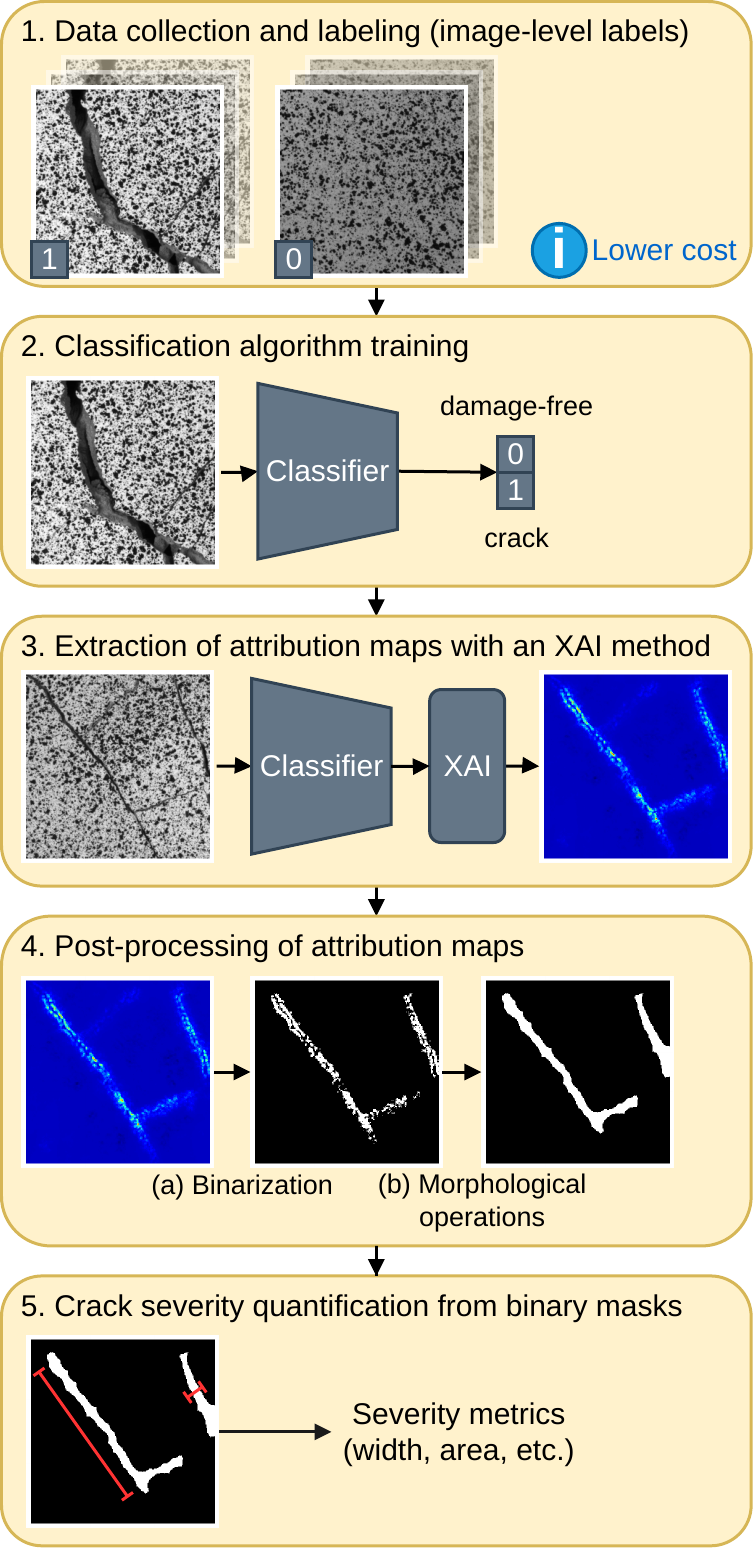}
        \caption{}
        \label{fig:workflow-xai}
     \end{subfigure}
    \caption{Workflows for automated image-based crack segmentation and severity quantification. (a) Supervised semantic segmentation workflow. (b) Workflow of the proposed weakly-supervised methodology based on XAI and classifier explanations. This methodology allows to generate approximate segmentation masks and quantify severity while circumventing the high cost of pixel-level labeling of training images.}
    \label{fig:methodology}
\end{figure}

The proposed methodology is illustrated in Figure~\ref{fig:workflow-xai} and compared to the standard supervised semantic segmentation workflow (Figure~\ref{fig:workflow-seg}). The data labeling cost of step 1 is order of magnitudes smaller compared to the pixel-level labeling required for training a supervised segmentation algorithm, as it only requires image-level labels. In step 2, we apply a convolutional neural network as the classifier. It is important to note that while our study focuses on the binary case, the methodology can be easily extended to handle multi-class and multi-label classification, accommodating  different types of damages that may co-occur in the same image. In the multi-class case, attribution maps can be extracted for each damage class. For step 3, we assume throughout this work that the crack segmentations are generated for unseen future images, which are represented as an independent hold-out test set. However, it is also possible to generate the segmentation masks for the training images. In practice, we observed no significant difference in the results. Therefore, we report the results solely based on the test set. After the post-processing (step 4), the resulting binary masks provide approximate segmentations of the cracks, allowing the quantification of crack severity in step 5.


%
\subsection{Explainable AI methods}

In this section, we present the explainable AI methods  evaluated in our study. We have included several popular XAI methods, each employing distinct approaches to compute relevance scores. All methods are able to generate attribution maps at the input resolution, which is a requirement to capture the thin structures of cracks. Moreover, these methods have easy-to-use available implementations, and a reasonable computational runtime. We intentionally avoided methods that output lower-resolution maps at the feature level, e.g., CAM \cite{zhou_learning_2016}, Grad-CAM \cite{selvaraju_grad-cam_2017} and heatmap network-based approaches with global pooling layers \cite{hwang_self-transfer_2016,dubost_gp-unet_2017,chatterjee_weakly-supervised_2022}. We also omitted very computationally expensive methods in this study, such as perturbation-based approaches.

\textbf{Input$\times$Gradient} \cite{baehrens_how_2010} is one of the earliest gradient-based explanation methods. It operates by computing the gradient for each input dimension at the current input value and then multiplying it with the input itself. This process reveals the change in the output resulting from an infinitesimal change of the input, thus indicating the local importance of each input dimension. However, its effectiveness is limited to the immediate local information provided by the gradient.

\textbf{Layer-wise Relevance Propagation (LRP)} \cite{bach_pixel-wise_2015,samek_layer-wise_2019} is a popular technique for decomposing a decision into pixel-wise relevance scores. It utilizes a backward propagation process and relies on access to the model internals such as weights and activation functions. LRP operates backward and propagates the relevance scores from the upper to the lower layers of the neural network, using specifically designed \textit{propagation rules} such as LRP-$0$, LRP-$\epsilon$, LRP-$\gamma$, LRP-$\alpha\beta$ and the $z^{\mathcal{B}}$-rule. For the best explanation quality, the rules are adjusted for different types of layers and activations. 

\textbf{Integrated Gradients} \cite{sundararajan_axiomatic_2017} calculates and accumulates the gradients along a straight-line path that interpolates between a reference input, called \emph{baseline}, and the input. The method satisfies two desirable properties known as completeness (i.e., the sum of the attributions over all features equals the difference between the model's output at the input and at the baseline) and implementation invariance (i.e., two networks with different implementations but the same outputs for all inputs produce identical attributions).

The \textbf{DeepLift} method \cite{shrikumar_learning_2019} computes relevance scores by comparing the network activations with a reference activation obtained on a baseline input. The contributions of each feature are computed as the differences between each neuron's activation and their reference activation, and propagated in a backward pass using a recursive algorithm similar to backpropagation. Being based on activations rather than gradients, it avoids shortcomings of gradient-based methods such as zero or discontinuous gradients. DeepLift's attribution quality is typically comparable to Integrated Gradients, but it runs significantly faster.

\textbf{DeepLiftShap} (also called DeepSHAP) is an application of DeepLift that uses SHAP (Shapley additive explanation) values as a measure of contribution \cite{lundberg_unified_2017}. Its attributions are estimated by sampling random images from a baseline distribution and averaging the resulting DeepLift attributions. Additionally, \textbf{GradientShap} approximates SHAP values by computing an expected gradient instead of an integral, as performed in Integrated Gradients, and can be seen as an approximation of the latter. Under model linearity and feature independence assumptions, SHAP values are approximated by the gradient expectation. The feature attributions are estimated by sampling random images from a baseline distribution and averaging their gradients multiplied by the difference between the input and the baseline.


The \textbf{Neural Network Explainer (NN-Explainer)} \cite{stalder_what_2022} is a recent method that trains an auxiliary network, referred to as the \emph{explainer}, to generate attribution maps for a trained classifier, referred to as the \emph{explanandum} (i.e., the model to be explained). These maps take the form of masks, denoted as $\mathbf{m} \in [0,1]^{W \times H}$, predicted for each class, where $W$ and $H$ represent the input image's width and height, respectively. Concretely, the explainer's architecture is similar to that of a segmentation network.The explainer is trained to minimize the cross-entropy of the explanandum within the image region selected by the mask and to maximize entropy (i.e., uncertainty) outside of the mask. The loss function also incorporates multiple regularization terms that penalize the area of the mask while encouraging smoothness. However, this formulation is valid for images containing one or multiple foreground objects and is not directly applicable in the context of damage classification, where one class represents the absence of foreground objects. To adapt the NN-Explainer to the context of crack detection, we propose a modification of the approach introduced in Section~\ref{sec:nn-explainer}.

The final method considered in this study, \textbf{B-cos networks} \cite{bohle_b-cos_2022}, is an explainable-by-design approach that aims at making the learned model inherently transparent. The authors propose to replace all linear transformations in the network, including convolutions, with a so-called \emph{B-cos transform}. This transform promotes alignment between weights and inputs during training, and demonstrates that the network can be faithfully summarized by a single input-dependent linear transformation. Attribution maps for a given class and input can then be obtained simply by visualizing the corresponding dimensions in the (input-dependent) matrix associated with this linear transform.

\subsection{Post-processing steps}

\begin{figure*}
    \centering
    \includegraphics[width=\textwidth]{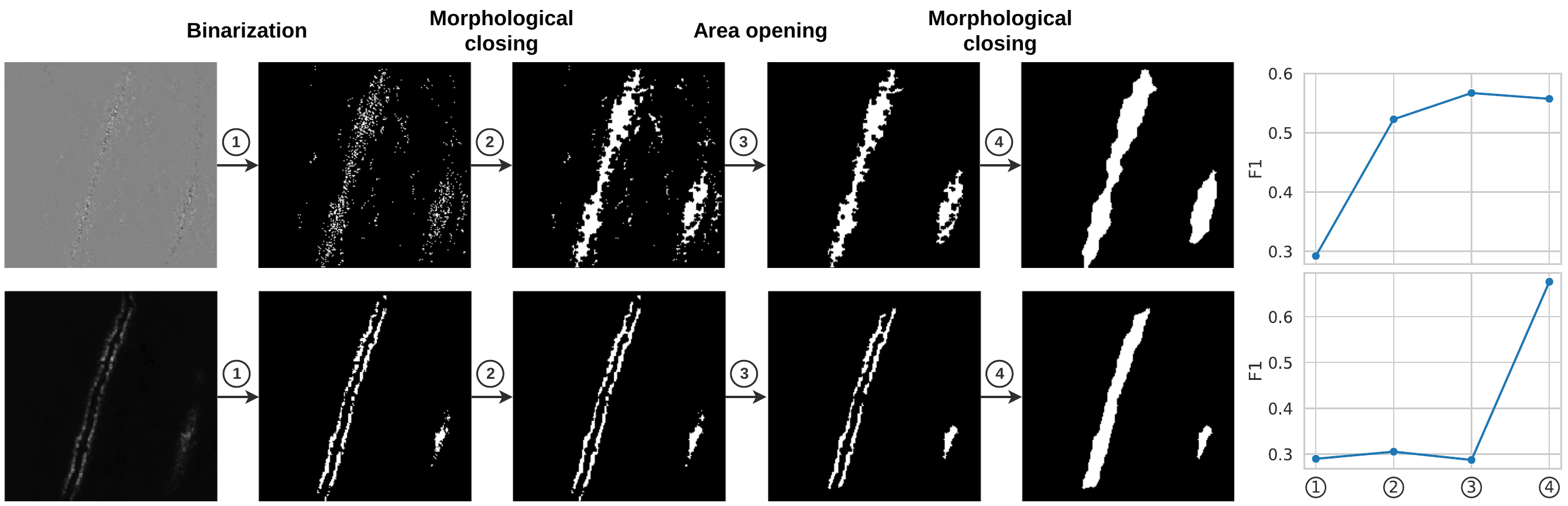}
    \caption{Illustration of the post-processing steps using Integrated Gradients (top) and LRP (bottom) attribution maps. (1) Binarization (2) First morphological closing (3) Area opening (4) Second closing. The curve on the right shows the evolution of F1 score at each post-processing step.}
    \label{fig:postproc}
\end{figure*}

The attribution maps undergo post-processing in two stages. In the first stage, we binarize the continuous attribution maps through thresholding. In the second stage, three morphological operations are applied as follows: (1) Closing, which involves dilation followed by erosion, to merge dense regions in the attribution maps; (2) Area opening with a minimum area threshold to remove remaining noise; (3) Second closing to close larger gaps in the resulting mask. Choosing an appropriate radius for the morphological closing is crucial, as a too large value will merge noisy attributions, while a too small value will result in holes in the mask. Generally, closing increases the mask area, thereby increasing recall with respect to the ground-truth segmentation. However, it may also introduce false positive pixels, thereby reducing the precision. The post-processing steps are visualized step-by-step in Figure~\ref{fig:postproc} for two different attribution methods (Integrated Gradients in the top row and LRP in the bottom row). Depending on the attribution method, different steps of the post-processing may or may not  improve the segmentation quality, as measured by the F1 score relative to the ground-truth segmentation.

Please note that the quality of the segmentation could potentially be further optimized by tuning the post-processing specifically for each attribution method and final application. However, for the sake of simplicity and to ensure a fair comparison, we used identical post-processing steps across all methods in our benchmark. Furthermore, it is important to mention that the evaluated explainability techniques and the post-processing steps do not incorporate specific knowledge about the structural characteristics of cracks, such as pixel connectivity or other effective regularization techniques used in the literature \cite{mei_cost_2020,mei_densely_2020,pantoja-rosero_topo-loss_2022}. While these techniques are not the focus of our study, they can be used in combination with the proposed methods to improve the resulting crack segmentation. This makes the proposed methodology general and applicable is various contexts, while enabling it to be further extended and improved for each specific application.

\subsection{Adaptation of the NN-Explainer for damage classification}\label{sec:nn-explainer}

\begin{figure}
    \centering
    \includegraphics[width=1.08\textwidth]{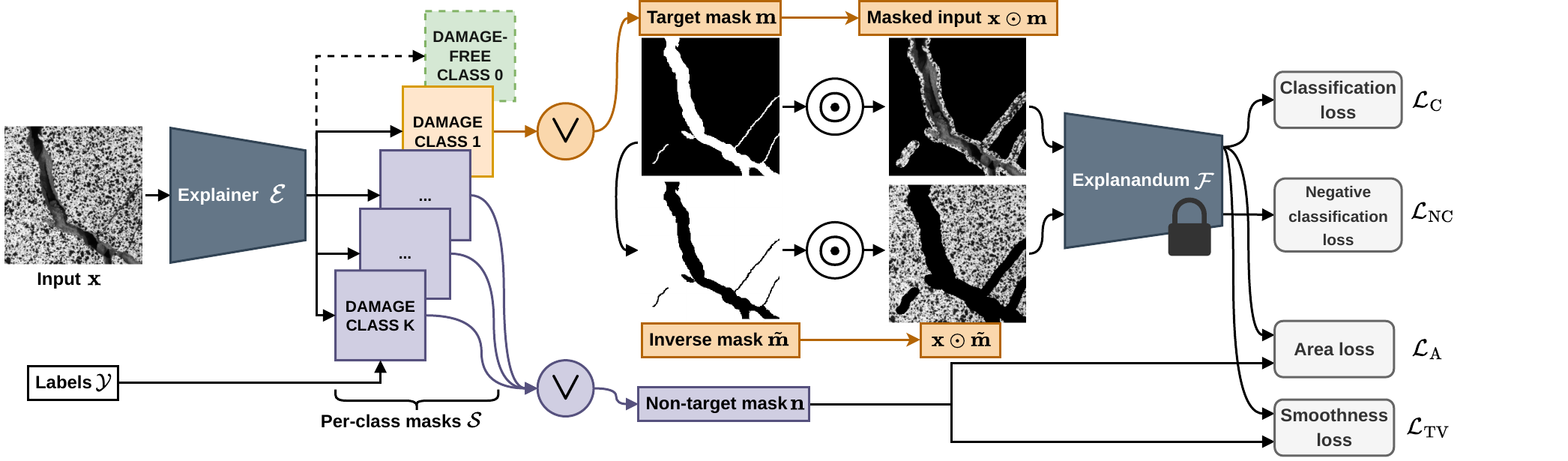}
    \caption{Overview of the NN-Explainer method \cite{stalder_what_2022} for damage classification with a negative (damage-free) class and $K$ positive (damage) classes. The \emph{explainer} $\mathcal{E}$ learns to predict masks $\mathcal{S}$ for each class. Class $0$ represents damage-free samples. Masks of positive classes present in the input are merged using their element-wise maximum (denoted by $\vee$ in the diagram) to create the target mask $\mathbf{m}$ (and its inverse $\tilde{\mathbf{m}}$), while masks of other positive classes that are not present form the non-target mask $\mathbf{n}$. In our application, there is a single damage class for cracks, and the non-target mask does not play a role. The \emph{explanandum} $\mathcal{F}$ (i.e., the model to be explained) is frozen.}
    \label{fig:damage-explainer}
\end{figure}
In this section, we derive our adaptation of the NN-Explainer \cite{stalder_what_2022} in the context of $(K+1)$-class multi-label classification, where $K$ represents the number of different damage types ranging from $1$ to $K$. In this setting, multiple damages can occur in the same image, while the negative class $0$ represents the damage-free class. The illustration of our method can be found in Figure~\ref{fig:damage-explainer}. The original NN-Explainer method proposes the following loss function to train the \emph{explainer} network:
\begin{equation}
    \mathcal{L}_{\mathcal{E}}(\mathbf{x}, \mathcal{Y}, \mathbf{m}, \mathbf{n}) = \mathcal{L}_{\text{C}}(\mathbf{x}, \mathcal{Y}, \mathbf{m}) + \lambda_{\text{E}} \mathcal{L}_{\text{E}}(\mathbf{x}, \tilde{\mathbf{m}}) + \lambda_{\text{A}} \mathcal{L}_{\text{A}}(\mathbf{m}, \mathbf{n}, \mathcal{S}) + \lambda_{\text{TV}} \mathcal{L}_{\text{TV}}(\mathbf{m}, \mathbf{n})
\end{equation}
where $\mathbf{x} \in \mathbb{R}^{C \times W \times H}$ represents the input image ($C$, $W$, $H$ denoting the image channels, width and height, respectively), $\mathcal{Y}$ is the set of positive target classes present in the image (defined at the image level), $\mathbf{m} \in [0,1]^{W \times H}$ is the aggregated mask generated for the target classes, $\tilde{\mathbf{m}} = 1 - \mathbf{m}$ is the inverse target mask, $\mathcal{S}$ is a set of per-class masks and $\mathbf{n} \in [0,1]^{W \times H}$ is the aggregated mask produced for non-target positive classes. The hyperparameters $\lambda_{\text{E}}$, $\lambda_{\text{A}}$ and $\lambda_{\text{TV}}$ are used for balancing the loss terms. Only positive samples containing at least one damage are used to train the explainer. In the remaining sections, we also adopt the notations from \cite{stalder_what_2022}, where $\mathcal{E}$ represents the explainer and $\mathcal{F}$ denotes the explanandum. We now discuss the different terms of the loss function and present our proposed modification.

\paragraph{Classification loss: $\mathcal{L}_{\text{C}}(\mathbf{x}, \mathcal{Y}, \mathbf{m})$} This loss function encourages the target mask $\mathbf{m}$ to highlight the regions in the input that are correctly classified by the trained model. It is computed as the sum of binary cross-entropies for each positive class present in the image, using the probabilities output by $\mathcal{F}$ when applied to the masked input: $\mathbf{p} = \mathcal{F}(\mathbf{x} \odot \mathbf{m})$, where $\odot$ represents element-wise multiplication. The expression for the classification loss is as follows:
\begin{equation*}
    \mathcal{L}_{\text{C}}(\mathbf{x}, \mathcal{Y}, \mathbf{m}) = -\frac{1}{K}\sum_{k=1}^{K} \llbracket k \in \mathcal{Y} \rrbracket \log (\mathbf{p}[k]) + \llbracket k \notin \mathcal{Y} \rrbracket \log (1 - \mathbf{p}[k]).
\end{equation*}
It is important to note that in the case of single-label classification, such as in our crack detection study, this loss is equivalent to the traditional cross-entropy.

\paragraph{Negative entropy loss: $\mathcal{L}_{\text{E}}(\mathbf{x}, \tilde{\mathbf{m}})$} We propose to modify this loss term in order to adapt it for damage detection. The original formulation of NN-Explainer \cite{stalder_what_2022} was designed for multi-label classification tasks where every input image contains one or several objects. In such cases, where there is no specific class for the absence of objects, NN-Explainer incorporates a loss term that maximizes the classification entropy (i.e., uncertainty) when the objects of interest are masked out, representing only background to the classifier. This is achieved by computing the negative entropy of the model probabilities on the inverse-masked input $\tilde{\mathbf{p}} = \mathcal{F}(\mathbf{x} \odot \tilde{\mathbf{m}})$ across all positive classes. The formulation for the negative entropy loss is as follows:
\begin{equation}
    \mathcal{L}_{\text{E}}(\mathbf{x}, \tilde{\mathbf{m}}) = \frac{1}{K}\sum_{k=1}^K \tilde{\mathbf{p}}[k] \log(\tilde{\mathbf{p}}[k]).
\end{equation}
However, in our case of damage detection, the classifier distinguishes between the presence and absence of objects (damages), with the absence (background) being represented by the negative (damage-free) class. In other words, the damage-free class is also a target class \emph{per se}. Therefore, we replace the entropy term with the cross-entropy against the negative class, as an input containing only damage-free regions should be classified as the negative class. Thus, we propose the negative classification loss $\mathcal{L}_{\text{NC}}$ as a replacement for the negative entropy loss:
\begin{equation}
    \mathcal{L}_{\text{NC}}(\mathbf{x}, \tilde{\mathbf{m}}) = -\log (\tilde{\mathbf{p}}[0]) - \frac{1}{K}\sum_{k=1}^{K} \log (1 - \tilde{\mathbf{p}}[k]).
\end{equation}

\paragraph{Area loss: $\mathcal{L}_{\text{A}}(\mathbf{m}, \mathbf{n}, \mathbf{S})$ and Smoothness loss: $\mathcal{L}_{\text{TV}}(\mathbf{m}, \mathbf{n})$} The area loss plays a crucial role in penalizing the size of the mask, ensuring that it remains as small as possible and preventing the trivial solution of a target mask with 1 values everywhere. It also sets constraints on the minimum and maximum allowed areas beyond which mask areas are penalized. The smoothness loss, based on the total variation, promotes smooth and artifact-free masks. No modifications have been made to these two losses from the original paper \cite{stalder_what_2022}.

Finally, our modified loss is expressed as follows:
\begin{equation}
    \mathcal{L}_{\mathcal{E}}(\mathbf{x}, \mathcal{Y}, \mathbf{m}, \mathbf{n}) = \mathcal{L}_{\text{C}}(\mathbf{x}, \mathbf{y}, \mathbf{m}) + \lambda_{\text{NC}} \mathcal{L}_{\text{NC}}(\mathbf{x}, \tilde{\mathbf{m}}) + \lambda_{\text{A}} \mathcal{L}_{\text{A}}(\mathbf{m}, \mathbf{n}, \mathbf{S}) + \lambda_{\text{TV}} \mathcal{L}_{\text{TV}}(\mathbf{m}, \mathbf{n}).
\end{equation}

\section{Experiments}\label{sec:experiments}

This section first presents the experimental data used in our experiments and the networks adopted for the crack classification task. We then provide details on the compared explainable AI (XAI) methods and the experimental settings. The following paragraphs report the results of our experiments and studies on the segmentation quality, crack severity quantification, and growth monitoring.

\subsection{Data}

We conducted experiments on the Experimental DIC (Digital Image Correlation) cracks dataset \cite{rezaie_comparison_2020,pantoja-rosero_topo-loss_2022}, which consists of 530 256$\times$256 image patches from stone masonry walls that were damaged in a shear-compression loading experiment conducted at the EESD laboratory at EPFL. The mm/pixel ratio is 0.43. All the images have annotated ground-truth segmentation masks for the cracked image patches. To perform the classification, we augmented this dataset with 874 additional negative patches taken from the same walls. 

\begin{figure}
    \centering
    \includegraphics[width=0.28\textwidth]{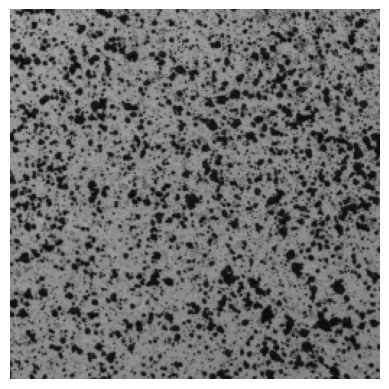}
    \includegraphics[width=0.28\textwidth]{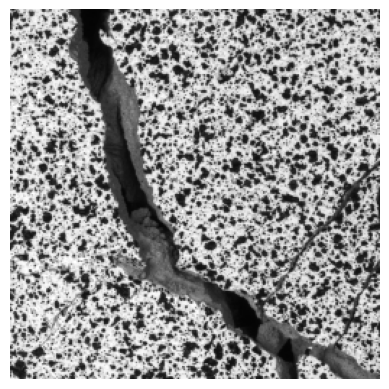}
    \includegraphics[width=0.28\textwidth]{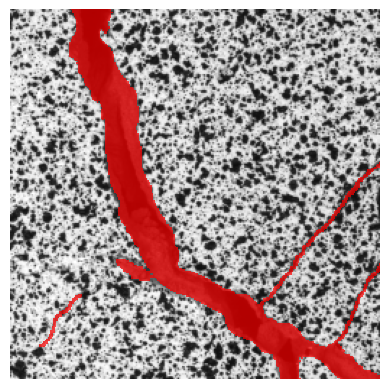}
    \caption{(left) Example negative (damage-free) image. (middle) Example positive (cracked) image. (right) Overlay of the ground-truth crack segmentation mask in red.}
    \label{fig:examples}
\end{figure}

The training and validation sets consist of 767 patches (301 positive/466 negative) and 328 patches (129 positive/199 negative), respectively. These patches were  extracted from 17 high-resolution images and randomly split. The test set, on which all results are reported, contains 309 patches (100 positive/209 negative) extracted from three different high-resolution images. Examples of images can be seen in Figure~\ref{fig:examples}. The complete dataset will be made available online, along with the code.

\subsection{Crack classifier}

The second step in our framework (see Figure~\ref{fig:workflow-xai} in Section~\ref{sec:method}) requires training a well-performing classifier to distinguish between positive and negative image patches. We use a VGG11 \cite{simonyan_very_2015} architecture with 128 neurons in the fully-connected layers, which we refer to as VGG11-128. We chose the VGG architecture because it is a widely-used standard CNN architecture in the related literature \cite{iglovikov_ternausnet_2018,pantoja-rosero_topo-loss_2022}, and it allows us to implement LRP rules \cite{seibold_explanations_2022} as well as B-cos networks \cite{bohle_b-cos_2022}.

To adapt the network to our small dataset, we reduced the number of neurons in the fully-connected layers from 4096 to 128 compared to the original VGG11. We trained the VGG11-128 model from scratch, using the cross-entropy loss and the Adam optimizer with a learning rate of $10^{-4}$, $\beta_1 = 0.9$, $\beta_2 = 0.999$ and a weight decay of $10^{-8}$. The only data augmentations applied are random horizontal and vertical flipping with probability $0.5$. We employed early stopping based on the validation fold to select the best-performing model. 

On the test set, the classifier achieved a balanced accuracy of 89\% , with a true positive rate (TPR) of 79\% and a true negative rate (TNR) of 99\%, as reported in Table~\ref{tab:classifier-results}. While the performance is sufficient to demonstrate the approach, it is worth noting that a more advanced classifier could potentially be trained by incorporating additional data augmentations.

\begin{table}
    \centering
    \caption{Crack classification performance on the DIC dataset (values in \%).}\label{tab:classifier-results}
    {\small
    \begin{tabular}{llccc}
    \toprule
     Model & Fold & Balanced accuracy & TPR & TNR \\
     \midrule
     \multirow{3}*{VGG11-128} & train & \phantom{0}99.9 & 100.0 & \phantom{0}99.8 \\
     & val & \phantom{0}96.2 & \phantom{0}95.3 & \phantom{0}97.0 \\
     & test & \phantom{0}89.0 & \phantom{0}79.0 & \phantom{0}99.0 \\
     \midrule
     \multirow{3}*{VGG11-128 B-cos} & train & \phantom{0}80.4 & \phantom{0}60.8 & 100.0 \\
     & val & \phantom{0}76.9 & \phantom{0}54.3 & \phantom{0}99.5 \\
     & test & \phantom{0}74.1 & \phantom{0}52.0 & \phantom{0}96.2 \\
     \bottomrule
    \end{tabular}
    }
\end{table}

For the B-cos network variant (VGG11-128 B-cos), we implemented the version with MaxOut units \cite{goodfellow_maxout_2013} and $B=2$, as recommended in \cite{bohle_b-cos_2022}. The training parameters are identical, except for the learning rate, which is reduced to $10^{-5}$. It is worth noting that the performance of the B-cos classifier model is significantly lower than that of the standard classifier model (VGG11-128). While the authors in \cite{bohle_b-cos_2022} observed only a small decrease  in performance on the CIFAR-10 benchmark, the performance gap is more pronounced in our task.



\rev{It is important to note that the choice of the VGG architecture does not restrict the methodology in any way. More recent architectures, such as residual networks (ResNet) and vision transformers (ViT), can also be used similarly. All the XAI methods used in our approach can be either directly applied to other architectures or with limited adaptations. In particular, gradient-based and activation-based methods can be directly applied, and an extension of LRP for Vision Transformers has been recently proposed in \cite{chefer_transformer_2021}. As we deal with a small-scale dataset throughout our study, more advanced architectures would not bring any benefit. We included additional results with different classifiers, namely ResNet-18 and ViT-B/16 (both initialized with ImageNet weights), in the Appendix. To conclude, the classifier architecture shall be adapted to the data size and complexity of the task, but the overall presented methodology remains applicable.}

\subsection{Compared methods and experimental settings}

In this study, we benchmark the ability of a total of eight different explainable AI methods to generate crack segmentation masks based on the pre-trained crack classifiers introduced in the previous section.

First, we evaluate six widely used post-hoc attribution methods for images, as introduced in Section~\ref{sec:method}): Input$\times$Gradient, Integrated Gradients (IntGrad), DeepLift, DeepLiftShap, GradientShap, and Layerwise Relevance Propagation (LRP). The model to be explained is the trained VGG11-128 classifier. We used the PyTorch implementations of the \texttt{captum} library \cite{kokhlikyan_captum_2020} for the first five methods, using the default parameters unless specified. For LRP, we adopt the segmentation rules proposed in \cite{seibold_explanations_2022}. The $z^{\mathcal{B}}$-rule is used for the first convolutional layer, the LRP-$\alpha\beta$ rule is used for the two lower convolutional layers, the LRP-$\gamma$ rule is used for all following convolutional layers, and the LRP-$\epsilon$ rule for all fully connected layers. The library \texttt{zennit} \cite{anders_software_2023} was used to implement these rules in our network.

The NN-Explainer method has been adapted to the setting with a negative class for damage-free samples, as explained in Section~\ref{sec:nn-explainer}. The explanandum is frozen and consists of the trained VGG11-128 classifier. For the explainer, a natural choice of architecture is the U-Net11 \cite{iglovikov_ternausnet_2018}, as it uses an encoder similar to the one in the classifier for feature extraction. The weighting hyperparameters in the loss function are set to $\lambda_{\text{NC}} = \lambda_{\text{A}} = \lambda_{\text{TV}} = 0.1$. The area loss also has two additional parameters to constrain the area range. We set the minimum and maximum values to $0.001$ and $0.15$, respectively, which roughly corresponds to the distribution of crack areas in the data, instead of the values $0.05$ and $0.3$ used in \cite{stalder_what_2022}. 

For the selection of class labels for the explainer training, we chose the ground-truth image-level training labels. As explained in \cite{stalder_what_2022}, it is more suited to handle attributions for false negative predictions. In the case of significant false positives, using the explanandum's predictions as labels might be better suited, but that is not the case here. We train the explainer model from scratch, using the Adam optimizer  with a learning rate of $10^{-5}$, $\beta_1 = 0.9$, $\beta_2 = 0.999$ and no weight decay.

For the B-cos network, we extract the visualizations by following the procedure described in the experimental section of the paper \cite{bohle_b-cos_2022}.

We also include unsupervised and supervised methods, not based on XAI, for comparison. 

The ``Raw method'' simply uses the raw pixel intensities, with the image only converted to grayscale before post-processing. We also evaluated different image enhancement techniques, such as Gaussian blurring, shading correction \cite{lee_technique_2007} and Min-Max Gray Level Discrimination \cite{hoang_detection_2018}, before the binarization. However, these methods did not improve the results, as they primarily address uneven illumination of the image and are unable to distinguish between the material patterns and the cracks in our dataset.
As a second unsupervised comparison method, we trained a convolutional autoencoder (CAE) to reconstruct only the damage-free training images, as done in \cite{chow_anomaly_2020}. Then, the pixel-wise reconstruction error is used to generate attribution maps for the testing images. The CAE has a VGG11 encoder followed by a fully-connected layer with 128 neurons and a 100-dimensional bottleneck, and a symmetric decoder. The mean squared error loss is used to train the CAE.

Finally, as a supervised oracle method, we trained a U-Net11 \cite{iglovikov_ternausnet_2018} on the ground-truth pixel-level segmentation labels of the training set. This serves as an upper bound on the performance that could be achieved by a supervised segmentation model with access to the fully labeled dataset at pixel-level. The U-Net was trained from scratch using the Dice loss for 100 epochs, with the Adam optimizer (learning rate $10^{-4}$, $\beta_1 = 0.9$, $\beta_2 = 0.999$). It is important to note that a fair comparison cannot be made between our methodology and the U-Net, as the U-Net relies on direct supervision from the pixel-level labels, which necessitates extensive annotation prior to training.

For the post-processing, we employ the \textit{simple} or \textit{GMM} thresholding strategies described in \cite{seibold_explanations_2022} for the binarization of attribution maps. In the second stage, we used an elliptical kernel for the morphological closing operations. The radius is set to $r=5\;\text{px}$ in the first closing. The area opening operation uses a minimum area of $50\;\text{px}^2$. Finally, the second closing uses a radius of $r=25\;\text{px}$ to close larger gaps in the masks. The parameters were fixed empirically and were kept identical across all methods in our benchmark, without specific tuning to enable a fair comparison.

\subsection{Importance of the baseline image}

\begin{figure}
    \centering
    \begin{subfigure}[c]{0.33\textwidth}
        \centering
         \includegraphics[width=0.8\textwidth]{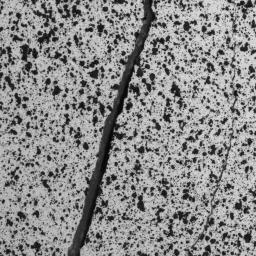}
         \caption{Input image}
     \end{subfigure}
    \begin{subfigure}[c]{0.66\columnwidth}
        \centering
         \includegraphics[width=0.4\columnwidth]{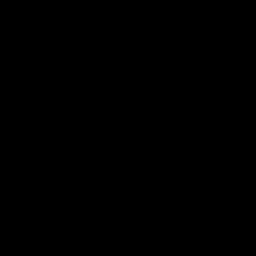}
         \includegraphics[width=0.4\columnwidth]{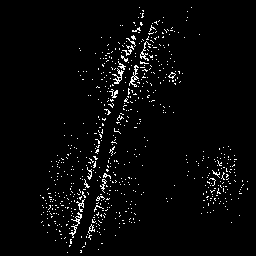}
         \caption{Zero baseline and resulting DeepLift attributions}
     \end{subfigure}
     \begin{subfigure}[c]{0.33\textwidth}
        \centering
         \includegraphics[width=0.8\textwidth]{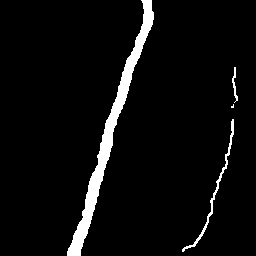}
         \caption{Ground-truth mask}
     \end{subfigure}
     \begin{subfigure}[c]{0.66\columnwidth}
        \centering
         \includegraphics[width=0.4\columnwidth]{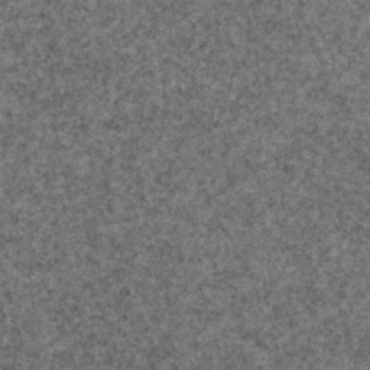}
         \includegraphics[width=0.4\columnwidth]{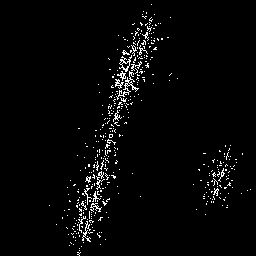}
         \caption{Mean damage-free baseline and resulting DeepLift attributions}
     \end{subfigure}
    \caption{Comparison between DeepLift attributions obtained with (b) the standard zero baseline and (d) our proposed baseline based on a sample of damage-free images. Attributions are more focused on the crack with our baseline. The behavior is similar for other attribution methods using a baseline image or distribution (e.g., Integrated Gradients).}
    \label{fig:deeplift-baseline-improvement}
\end{figure}

Some relevance methods require a baseline image (such as IntGrad and DeepLift) or a baseline image distribution (such as DeepLiftShap and GradientShap) as a reference point for computing changes or gradients compared to the input. The commonly used standard baselines for these methods are a zero matrix (i.e., a black image) or a random normal baseline distribution. However, we observed that with these standard baselines the methods performed poorly, as the attributions tend to be spread over large areas surrounding the crack. To address this issue, we propose an alternative approach of sampling a set of images from the damage-free class (without cracks) as baselines. For  methods like IntGrad and DeepLift, we use the mean of the sampled damage-free images as the baseline. For DeepLiftShap and GradientShap, we use these damage-free images as the baseline distribution, with 10 samples in order to keep the runtimes reasonable. This modification significantly improves the quality of attributions, as the focus of the attribution shifts more towards the crack itself rather than the healthy regions of the image. An example illustrating this improvement is visualized in  Figure~\ref{fig:deeplift-baseline-improvement}. Quantitatively, when using the standard baseline, IntGrad obtained an F1-score of only 11.13\% (after simple thresholding), whereas it increased to 18.91\% when using the mean damage-free image baseline (see next paragraph for complete results). All results presented in this paper for IntGrad, DeepLift, DeepLiftShap, and GradientShap employ our proposed baseline method.

\section{Results}\label{sec:results}

%
\subsection{Segmentation quality evaluation}

This paragraph presents the results of our benchmark in terms of the segmentation quality of the generated masks. While localization performance is commonly used to evaluate attribution methods in a quantitative way, such as the grid pointing game \cite{zhang_top-down_2018,bohle_convolutional_2021,bohle_b-cos_2022}, segmentation quality provides a more fine-grained measure of localization performance. However, evaluating segmentation quality requires access to ground-truth segmentation masks \cite{seibold_explanations_2022,stalder_what_2022}. Fortunately, in our case, we have access to the true segmentation masks of the DIC images, allowing us to evaluate the segmentation quality metrics such as F1 score (also known as Dice score), Precision, Recall and Intersection-over-Union (IoU or Jaccard index) on the test set. These metrics are calculated based on the per-pixel true positives (TP), false positives (FP) and false negatives (FN) in relation to the ground-truth segmentation mask:

\begin{align*}
    &\text{Precision} = \frac{\text{TP}}{\text{TP} + \text{FP}} \hspace{1.5cm}
    \text{Recall} = \frac{\text{TP}}{\text{TP} + \text{FN}}\\
    &\text{F1} = \frac{2\times\text{TP}}{2\times\text{TP} + \text{FP} + \text{FN}} \hspace{1cm}
    \text{IoU} = \frac{\text{TP}}{\text{TP} + \text{FP} + \text{FN}}\\
\end{align*}

For each evaluated method, we provide the results using various combinations of thresholding and morphological post-processing, as outlined in the methodology proposed in Figure~\ref{fig:workflow-xai} in Section~\ref{sec:method}. In Table~\ref{tab:segmentation-results}, we present the quantitative findings. Methods are grouped by type of supervision: XAI-based (weakly-supervised, following the proposed methodology), unsupervised, and fully supervised.

The two best-performing XAI methods are DeepLiftShap and LRP, achieving F1/IoU scores of 38.1\%/23.60\% and 37.43\%/23.03\%, respectively, when used in conjunction with simple thresholding and morphological post-processing. The next two best methods are DeepLift and the B-cos network. Although these scores may appear relatively low, particularly when compared to the performance of the supervised U-Net model (83.67\%/71.93\%), it is important to note that these segmentations were generated  without explicit supervision, and only required a trained binary classifier for cracked and non-cracked image patches. Unlike the U-Net oracle model, no pixel-level labels were necessary. The unsupervised approaches, using raw pixels and the CAE, both obtain very low performance (around 5\% F1 after post-processing). In both cases, the resulting masks comprise most of the dark pixels in the images, without separating the crack from the material texture. The noisy textures of the images in our dataset, common for construction materials, hinders the autoencoder from producing accurate reconstructions. Since there is no supervisory signal, the CAE cannot separate the discriminative signal from noise and only reconstructs the mean of the image distribution. While the CAE performs well on less noisy images as shown in \cite{chow_anomaly_2020}, it struggles with more complex cases like ours.
\begin{table}
    \setlength{\aboverulesep}{0pt}
    \setlength{\belowrulesep}{0pt}
    \centering
    \caption{Crack segmentation quality obtained with the proposed weakly-supervised methodology using different explainable AI methods and post-processing techniques (values in \%). Best and second-best scores in bold underlined and bold, respectively.}
    \label{tab:segmentation-results}
    {\footnotesize
    \begin{tabular}{|l|l|cc|cccc|}
        \toprule
        \multicolumn{2}{|c|}{\multirow{2}*{Method}} & \multicolumn{2}{c|}{Post-processing} & \multirow{2}*{F1} & \multirow{2}*{Precision} & \multirow{2}*{Recall} & \multirow{2}*{IoU} \\
        \multicolumn{2}{|c|}{} & thresh. & morph. & & & & \\
        \midrule
        \multirow{32}*{\rotatebox[origin=c]{90}{XAI-based (weakly-supervised)}}
        & \multirow{4}*{Input$\times$Gradient} & simple & \xmark & 14.64 & 14.44 & 14.85 & \phantom{0}7.90 \\
        & & simple & \cmark & 23.30 & 14.37 & 61.55 & 13.19 \\
        & & GMM & \xmark & 21.54 & 16.17 & 32.28 & 12.07 \\
        & & GMM & \cmark & 20.76 & 12.41 & 63.52 & 11.58 \\
        \cmidrule{2-8}
        & \multirow{4}*{IntGrad} & simple & \xmark & 18.91 & 26.63 & 14.66 & 10.44 \\
        & & simple & \cmark & 27.74 & 20.81 & 41.56 & 16.10 \\
        & & GMM & \xmark & 28.92 & 25.71 & 33.06 & 16.91 \\
        & & GMM & \cmark & 25.96 & 17.02 & 54.68 & 14.92 \\
        \cmidrule{2-8}
        & \multirow{4}*{DeepLift} & simple & \xmark & 22.58 & 34.97 & 16.67 & 12.73 \\
        & & simple & \cmark & 34.44 & 28.75 & 42.96 & 20.81 \\
        & & GMM & \xmark  & 31.70 & 27.06 & 38.27 & 18.84 \\
        & & GMM & \cmark & 29.55 & 19.81 & 58.12 & 17.33 \\
        \cmidrule{2-8}
        & \multirow{4}*{DeepLiftShap} & simple & \xmark & 24.03 & \underline{\textbf{47.07}} & 16.14 & 13.66 \\
        & & simple & \cmark & \underline{\textbf{38.19}} & \textbf{36.37} & 40.21 & \underline{\textbf{23.60}} \\
        & & GMM & \xmark & 37.07 & 34.25 & 40.39 & 22.75 \\
        & & GMM & \cmark & 33.10 & 23.11 & 58.32 & 19.83 \\
        \cmidrule{2-8}
        & \multirow{4}*{GradientShap} & simple & \xmark  & 13.88 & 16.39 & 12.04 & \phantom{0}7.46 \\
        & & simple & \cmark & 20.61 & 14.09 & 38.38 & 11.49 \\
        & & GMM & \xmark & 19.78 & 20.02 & 19.55 & 10.97 \\
        & & GMM & \cmark & 21.27 & 15.32 & 34.79 & 11.90 \\
        \cmidrule{2-8}
        & \multirow{4}*{LRP} & simple & \xmark & 22.17 & 27.28 & 18.67 & 12.46 \\
        & & simple & \cmark & \textbf{37.43} & 35.06 & 40.16 & \textbf{23.03} \\
        & & GMM & \xmark & 28.39 & 21.83 & 40.57 & 16.54 \\
        & & GMM & \cmark & 36.16 & 25.84 & 60.15 & 22.07 \\
        \cmidrule{2-8}
        & \multirow{4}*{NN-Explainer} & simple & \xmark & 25.17 & 14.94 & \textbf{79.80} & 14.40 \\
        & & simple & \cmark & 24.83 & 14.65 &  \underline{\textbf{81.33}} & 14.17 \\
        & & GMM & \xmark & 29.17 & 18.18 & 73.78 & 17.07 \\
        & & GMM & \cmark & 28.92 & 17.88 & 75.62 & 16.91 \\
        \cmidrule{2-8}
        & \multirow{4}*{B-cos network} & simple & \xmark & 23.31 & 16.87 & 37.69 & 13.19 \\
        & & simple & \cmark & 16.57 & 9.57 & 61.60 & \phantom{0}9.03 \\
        & & GMM & \xmark & 31.35 & 24.54 & 43.40 & 18.59 \\
        & & GMM & \cmark & 30.68 & 20.40 & 61.81 & 18.12 \\
        \midrule
        \multirow{8}*{\rotatebox[origin=c]{90}{Unsupervised}}
        & \multirow{4}*{Raw} & simple & \xmark & 12.18 & \phantom{0}6.54 & 89.46 & \phantom{0}6.49 \\
        & & simple & \cmark & \phantom{0}4.73 & \phantom{0}2.42 & 100.0 & \phantom{0}2.42 \\
        & & GMM & \xmark & 18.05 & 10.21 & 78.00 & \phantom{0}9.92 \\
        & & GMM & \cmark & \phantom{0}7.88 & \phantom{0}4.12 & 91.36 & \phantom{0}4.10 \\
        \cmidrule{2-8}
        & \multirow{4}*{CAE} & simple & \xmark & 12.39 & \phantom{0}7.05 & 51.22 & \phantom{0}6.60 \\
        & & simple & \cmark & \phantom{0}5.93 & \phantom{0}3.07 & 90.09 & \phantom{0}3.06 \\
        & & GMM & \xmark & 15.36 & \phantom{0}9.39 & 42.11 & \phantom{0}8.32 \\
        & & GMM & \cmark & 13.32 & \phantom{0}7.57 & 55.68 & \phantom{0}7.14 \\
        \midrule
        \multicolumn{4}{|l|}{Supervised U-Net (oracle)} & 83.67 & 82.22 & 85.17 & 71.93 \\
        \bottomrule
    \end{tabular}
    }
\end{table}
In terms of post-processing, the GMM thresholding strategy generally yields the best performance. However, when morphological operations are applied, the simple strategy produces superior results. The selection of the post-processing techniques, particularly the choice of morphological operations, involves  a trade-off between precision, recall and the visual aspect of the resulting mask (e.g., presence of noise or holes in the mask). Therefore, the preference for specific post-processing strategies should be based on the end user's preference and the requirements of downstream tasks.

\begin{figure}
    \centering
    \begin{subfigure}[c]{\textwidth}
        \centering
        \includegraphics[width=\textwidth]{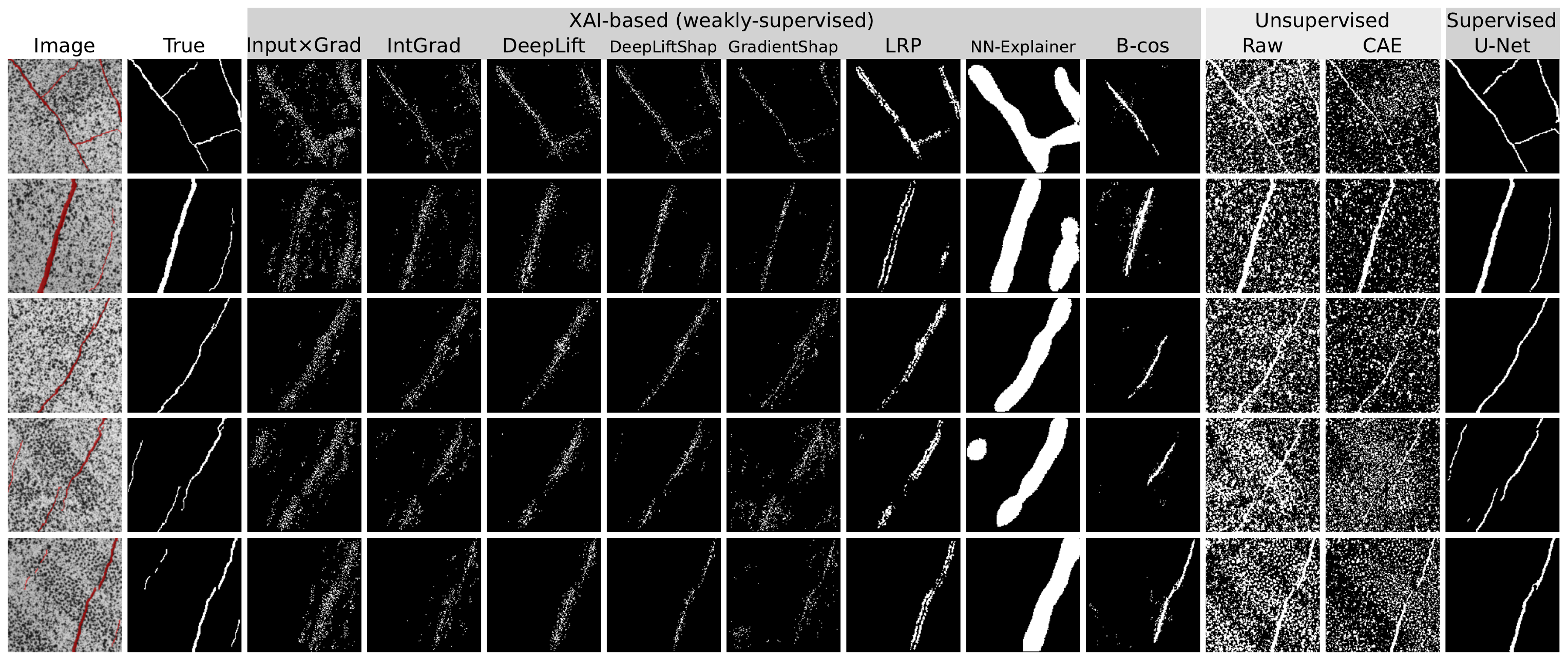}
        \caption{}
        \label{fig:maps-simple}
    \end{subfigure}
    \begin{subfigure}[c]{\textwidth}
        \centering
        \includegraphics[width=\textwidth]{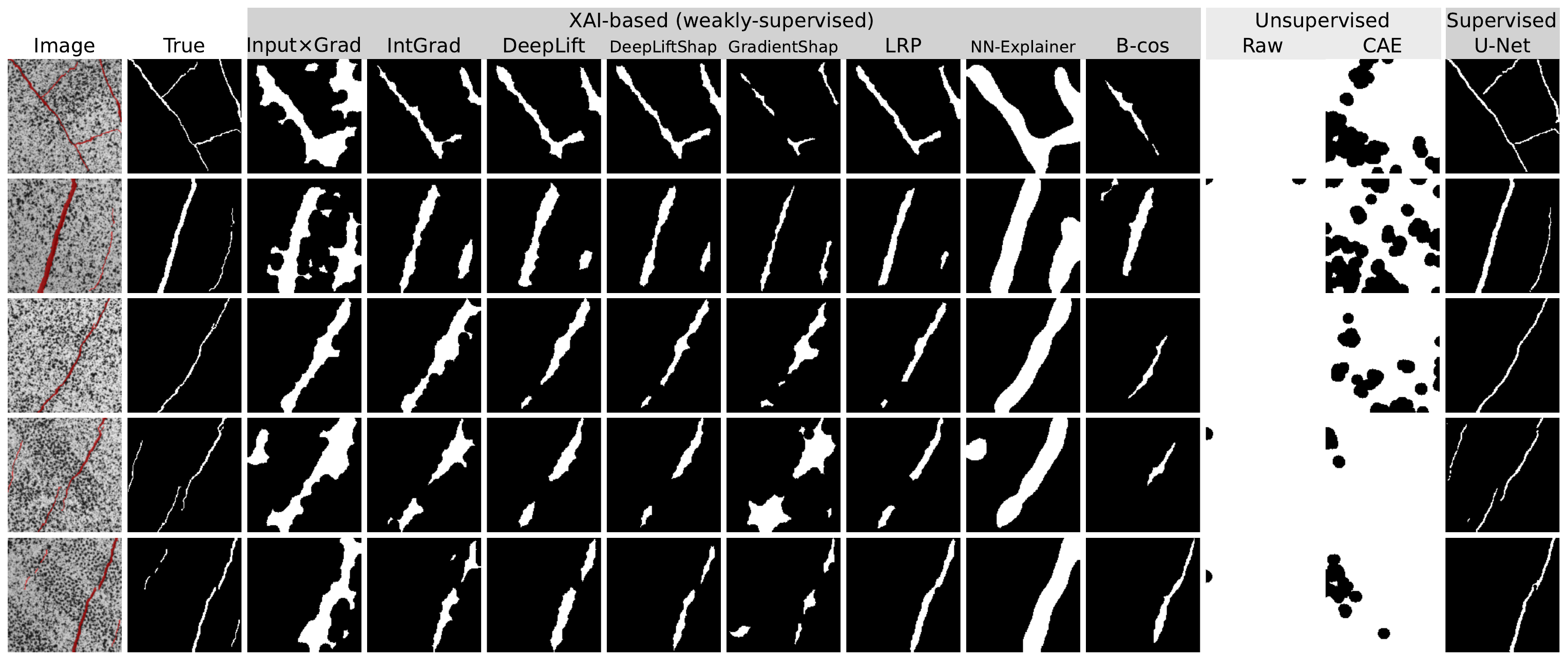}
        \caption{}
        \label{fig:maps-simple-final}
     \end{subfigure}
     \caption{Visualization of crack segmentation masks obtained through different explainable AI methods (a) after binarization of the attribution maps, and (b) after full post-processing with morphological operations. Ground-truth masks are highlighted in red on the original images. U-Net requires supervised training with pixel-level labels and serves as a reference (oracle).}
    \label{fig:maps}
\end{figure}

Visualizations for five examples are shown in Figure~\ref{fig:maps}, depicting the results after binarization (with the simple thresholding strategy) in Figure~\ref{fig:maps-simple}, and after the application of morphological operations in  Figure~\ref{fig:maps-simple-final}. The figure displays the image and ground-truth mask in the first two columns, followed by masks obtained by XAI methods, unsupervised methods, and the supervised U-Net. In terms of qualitative assessment, DeepLift, DeepLiftShap, and LRP produce visualizations that closely resemble the ground-truth segmentation. LRP and the B-cos network yield cleaner masks with less noise, although the attributions of the B-cos network are incomplete. Other methods exhibit more scattered and noisy attributions, resulting in masks that are too spread out. This qualitative observation corroborates with the quantitative results, showing higher recall but lower precision. The main source of error lies in the omission of very thin cracks when multiple cracks are present in the image (as observed in examples 4 and 5 in Figure~\ref{fig:maps-simple-final}). In such cases, the attributions of the thinner cracks are overshadowed by noise and are lost during the post-processing stage. It is worth noting that even the supervised U-Net fails to capture the two thinner cracks in the last example. The NN-Explainer method shows promising results but produces masks that are too wide, resulting in a high recall but low precision. We believe this may be due to two main reasons. Firstly, the formulation of the explainer's training loss fails to effectively penalize the mask area, making it difficult to tightly fit around the crack. The mean total area penalty used  in the loss formulation is not suitable for capturing thin, skeleton-like structures such as cracks. Secondly, masking out image regions with zero values (i.e., black pixels) may not be ideal for cracks that are comprised of dark pixels. An \emph{in-distribution} masking operation might be more appropriate and could be explored in future work. Lastly, B-cos networks also show promising results but suffer from partial coverage of attributions over the object, failing to highlight the full length of the cracks and thereby compromising the segmentation quality. B-cos explanations are also constrained by the inferior classification performance of the B-cos network. As explained previously, the unsupervised methods are unable to separate the crack from the background texture, resulting in masks that cover the entire image.

\subsection{Augmentation smoothing}

Augmentation smoothing (AugSmooth) is a technique that involves averaging multiple attribution maps obtained using Test-Time Augmentations (TTA). Originally introduced with Grad-CAM \cite{selvaraju_grad-cam_2017}, AugSmooth improves localization at the expense of the additional computation incurred by TTA, requiring to predict and extract attribution maps for each augmented input \cite{gildenblat_pytorch_2021}. In our methodology, AugSmooth is applied during the generation of attribution maps, prior to post-processing. We conducted experiments using LRP+AugSmooth, and the results are presented in Table~\ref{tab:tta-results}. For TTA, we used 6 combinations of random horizontal and vertical flipping, as well as random intensity scaling by factors $0.9$, $1.0$ or $1.1$, following the approach described in \cite{gildenblat_pytorch_2021}. When combined with simple thresholding, substantial improvements were observed, with an absolute increase of +2.01\% in F1 score, primarily attributed to an increased recall (+4.22\%). However, it is worth noting that LRP+AugSmooth slightly degraded results (-0.35\%) when used in conjunction with GMM thresholding, which already exhibited high recall, due to a decrease in precision. Ultimately, when incorporating post-processing, LRP+AugSmooth achieved an F1 score approaching 40\%.

\begin{table}
    \setlength{\aboverulesep}{0pt}
    \setlength{\belowrulesep}{0pt}
    \centering
    \caption{Crack segmentation quality obtained using augmentation smoothing (AugSmooth) with Test-Time Augmentations for LRP (values in \%), with the delta compared to without AugSmooth. Best score in bold.}
    \label{tab:tta-results}
    {\small
    \begin{tabular}{|c|cc|cccc|}
        \toprule
        \multirow{2}*{Method} & \multicolumn{2}{c|}{Post-processing} & \multirow{2}*{F1} & \multirow{2}*{Precision} & \multirow{2}*{Recall} & \multirow{2}*{IoU} \\
        & thresh. & morph. & & & & \\
        \midrule
        & simple & \xmark & 24.21 {\scriptsize\textcolor{lightblue}{+2.04}} & 27.45 {\scriptsize\textcolor{lightblue}{+0.17}} & 21.65 {\scriptsize\textcolor{lightblue}{+2.98}} & 13.77 {\scriptsize\textcolor{lightblue}{+1.31}} \\
         LRP + & simple & \cmark & \textbf{39.44} {\scriptsize\textcolor{lightblue}{+2.01}} & \textbf{35.49} {\scriptsize\textcolor{lightblue}{+0.43}} & 44.38 {\scriptsize\textcolor{lightblue}{+4.22}} & \textbf{24.57} {\scriptsize\textcolor{lightblue}{+1.54}} \\
         AugSmooth & GMM & \xmark & 29.09 {\scriptsize\textcolor{lightblue}{+0.70}} & 21.41 {\scriptsize\textcolor{red}{-0.42}} & 45.38 {\scriptsize\textcolor{lightblue}{+4.81}} & 17.02 {\scriptsize\textcolor{lightblue}{+0.48}} \\
         & GMM & \cmark & 35.81 {\scriptsize\textcolor{red}{-0.35}} & 24.91 {\scriptsize\textcolor{red}{-0.93}} & \textbf{63.68} {\scriptsize\textcolor{lightblue}{+3.53}} & 21.81 {\scriptsize\textcolor{red}{-0.26}} \\
        \bottomrule   
    \end{tabular}
    }
\end{table}

\subsection{Crack severity quantification}

To evaluate the severity of the damage, we calculated the number of cracks per patch (CPP) \cite{pantoja-rosero_topo-loss_2022}, the total crack area per patch, and the maximum crack width. The width estimation method from \cite{carrasco_image-based_2021} was used to estimate the maximum crack width. In Table~\ref{tab:severity-results}, we report the mean absolute error (MAE) and mean absolute percentage error (MAPE, in \%) of each method compared to the corresponding ground-truth metric extracted from the ground-truth mask. Results for the Raw and CAE methods have been omitted as the resulting masks do not allow meaningful estimation of crack severity.
\begin{table}
    \setlength{\aboverulesep}{0pt}
    \setlength{\belowrulesep}{0pt}
    \centering
    \caption{Assessment of crack severity estimation using number of cracks per patch (CPP), crack area per patch and maximum crack width. The table reports the mean absolute error (MAE) or mean absolute percentage error (MAPE, in \%) with the ground-truth severity measure (lower is better). Best and second-best scores in bold underlined and bold, respectively.}
    \label{tab:severity-results}
    {\footnotesize
    \begin{tabular}{|l|l|cc|ccc|}
        \toprule
        \multicolumn{2}{|c|}{\multirow{2}*{Method}} & \multicolumn{2}{c|}{Post-processing} & CPP & Area & Width \\
        \multicolumn{2}{|c|}{} & thresh. & morph. & MAE & MAPE & MAPE (MAE px$\vert$mm) \\
        \midrule
        \multirow{16}*{\rotatebox[origin=c]{90}{XAI-based (weakly-supervised)}}
        & \multirow{2}*{Input$\times$Gradient} 
        & simple & \cmark & 1.13 & 448.1 & 358.8 (27.04$\vert$11.63) \\
        & 
        & GMM & \cmark & 1.14 & 629.3 & 404.4 (29.08$\vert$12.50) \\
        \cmidrule{2-7}
        & \multirow{2}*{IntGrad} 
        & simple & \cmark & 0.94 & 271.3 & 268.9 (18.75$\vert$8.06) \\
        & 
        & GMM & \cmark & 1.10 & 467.3 & 352.3 (25.82$\vert$11.11) \\
        \cmidrule{2-7}
        & \multirow{2}*{DeepLift} 
        & simple & \cmark & 0.81 & 146.0 & 264.6 (18.78$\vert$8.08) \\
        & 
        & GMM & \cmark & \underline{\textbf{0.72}} & 352.1 & 374.2 (27.77$\vert$11.94) \\
        \cmidrule{2-7}
        & \multirow{2}*{DeepLiftShap} 
        & simple & \cmark & \textbf{0.78} & \textbf{103.6} & \textbf{189.2 (12.67$\vert$5.45)} \\
        & 
        & GMM & \cmark & 0.82 & 310.7 & 344.6 (23.84$\vert$10.25) \\
        \cmidrule{2-7}
        & \multirow{2}*{GradientShap} 
        & simple & \cmark & 1.76 & 338.8 & 295.5 (21.52$\vert$9.25) \\
        & 
        & GMM & \cmark & 1.71 & 317.9 & 343.6 (24.43$\vert$10.50) \\
        \cmidrule{2-7}
        & \multirow{2}*{LRP} 
        & simple & \cmark & 0.90 & \phantom{0}\underline{\textbf{91.0}} & \underline{\textbf{163.1 (11.86$\vert$5.10)}} \\
        & 
        & GMM & \cmark & 0.82 & 261.8 & 257.6 (18.20$\vert$7.83) \\
        \cmidrule{2-7}
        & \multirow{2}*{NN-Explainer} 
        & simple & \cmark & 0.82 & 716.1 & 430.4 (31.41$\vert$13.51) \\
        & 
        & GMM & \cmark & 0.89 & 600.5 & 411.8 (29.04$\vert$12.49) \\
        \cmidrule{2-7}
        & \multirow{2}*{B-cos network} 
        & simple & \cmark & 1.77 & 1325.9 & 195.2 (17.87$\vert$7.68) \\
        & 
        & GMM & \cmark & 4.04 & 392.4 & 239.7 (23.04$\vert$9.91) \\
        \midrule
        \multicolumn{4}{|l|}{Supervised U-Net (oracle)} & 0.74 & \phantom{0}20.1 & \phantom{0}20.8 (1.67$\vert$0.72) \\
        \bottomrule
    \end{tabular}
    }
\end{table}
When estimating the number of CPP, most methods achieve a mean absolute error of less than 1, which is close to the performance of the U-Net supervised oracle model (0.74). The two best-performing methods are DeepLift (0.72) and DeepLiftShap (0.78), followed by LRP and NN-Explainer. However, all methods exhibit high errors in estimating crack area and width. This is primarily due to the methods overestimating the extent of the cracks or missing some cracks, as observed in Figure~\ref{fig:maps-simple-final}. The choice of post-processing, which favors clean masks but increases their size, also contributes to this issue. Specifically, the simple thresholding is more suitable compared to the GMM strategy, which tends to overestimate the crack size. The LRP method provides the most accurate assessment of crack area and width. On average, the error is 91\% for crack area, and 163\% for maximum crack width, which corresponds to a 2X to 3X range. Although these errors may seem high, it is important to consider that we are dealing with very thin structures that typically cover about 1\% of the image area, and mistakes of a few pixels result in a high percentage of error. In conclusion, our XAI-based methodology offers a rough estimate of crack severity metrics, but it is not as accurate as a fully supervised segmentation approach (the U-Net obtains 20\% MAPE on both area and width). However, it is worth noting that for tasks involving extracting the skeleton from the binary mask, the overestimation of the crack width is not a critical issue. Furthermore, if this overestimation is consistent, results can be calibrated and still be valuable for crack growth monitoring, which we study in the following section.

\subsection{Crack growth monitoring}

In this section, we investigate the slightly different task of crack growth monitoring, which involves studying the evolution of crack size or severity over time. This task is equally, if not more, important, than accurately estimating severity metrics. If we observe the same damage at different time points, our methodology would be valuable if it can also highlight a potential difference in estimated severity, even if the absolute value may not be very precise. Since obtaining real images showing the growth of cracks over time is challenging, we have designed an artificial experiment as a proof-of-concept.

\begin{figure}
    \centering
    \includegraphics[width=0.8\textwidth]{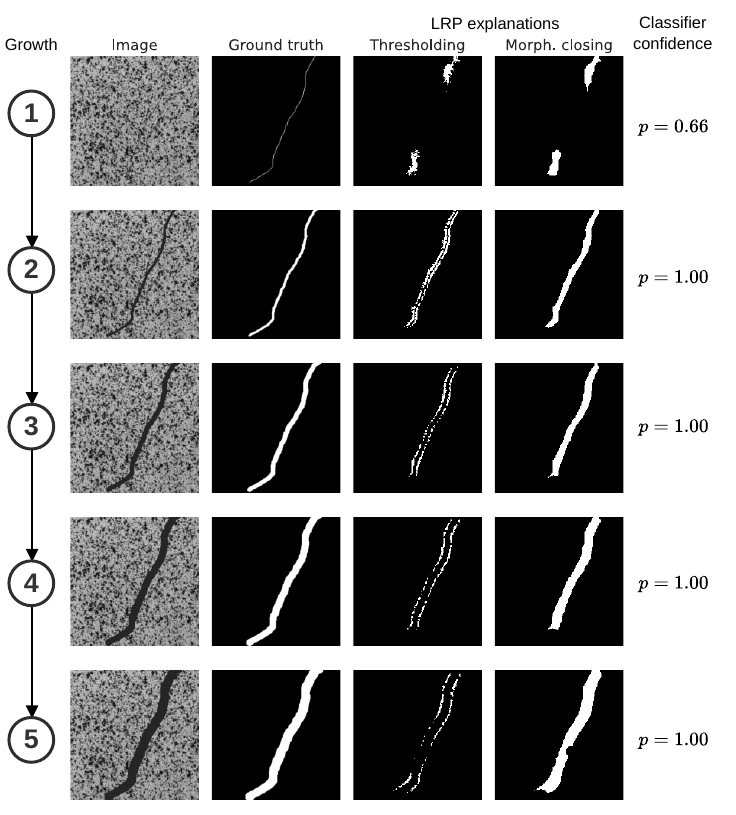}
    \caption{Visualization of artificial linear crack growth trajectories for crack growth monitoring. The generated images and their corresponding ground-truth masks are displayed in the first two columns. LRP explanations are displayed after thresholding (third column), and after final post-processing with morphological operations (last column). In the first image, the crack is too thin and the classifier has a low prediction confidence (66\% softmax score), explaining the poor corresponding explanation.}
    \label{fig:crack-growth-images}
\end{figure}

For this experiment, we randomly sample 100 damage-free images and 100 crack masks from the DIC dataset. For each pair of samples, we simulate a linear growth trajectory by generating a sequence of five images using the original crack skeleton, growing it linearly by repeatedly applying a dilation operation (using an elliptical kernel with radius $r=5\;\text{px}$), and overlaying the grown crack onto the damage-free image. We would like to mention that this experiment does not cover the crack branching during their growth, or the initiation of new cracks. We then follow the methodology outlined in Figure~\ref{fig:methodology}, using the same pre-trained classifier as in previous experiments. We derive the segmentation masks from the attribution maps of the positive class for each XAI method compared in this study. Finally, we calculate the crack areas and maximum widths based on the true and estimated masks in each growth trajectory.

One example of a generated growth trajectory is illustrated in Figure~\ref{fig:crack-growth-images}. The generated images and their corresponding ground-truth masks are displayed in the first two columns. LRP explanations are displayed after thresholding (third column), and after final post-processing with morphological operations (last column). Visually, we observe that the resulting mask accurately follows the growth of the crack, except for the first image in the sequence. In this case, the crack was too thin and the classifier exhibited low confidence in its prediction ($0.66$ softmax probability score for the crack class), resulting in a poor corresponding explanation. For comparison, the softmax confidence scores were equal to $1.00$ for the other images in the sequence. 

\begin{figure}
    \centering
    \includegraphics[width=\textwidth]{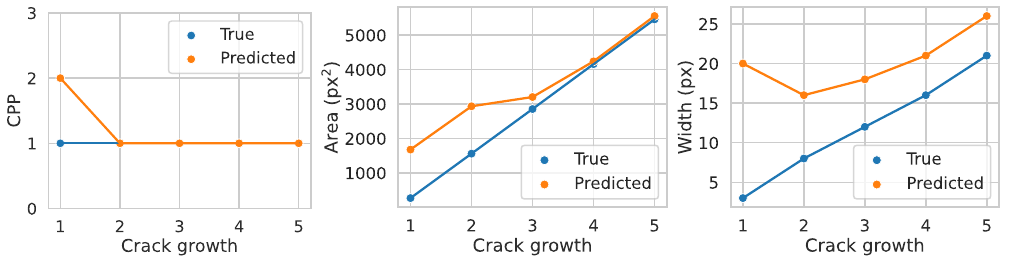}
    \caption{Evolution of true and estimated severity metrics as a function of crack growth, using our methodology with LRP, simple thresholding and morphological closing. Even if the area and width are over-estimated, they allow to monitor the growth of the crack, as long as the classifier's explanation is relevant. The classifier exhibited low prediction confidence in the first growth step, leading to a poor explanation and severity quantification.}
    \label{fig:crack-growth-metrics}
\end{figure}

On Figure~\ref{fig:crack-growth-metrics}, we illustrate the evolution of true and estimated severity metrics as crack growth progresses. With the exception of image 1, the metrics consistently exhibit a monotonically increasing trend and closely align with the true values, particularly as the crack becomes larger. While there is some overestimation in both area and width, they still provide effective means for monitoring crack growth, as long as the classifier’s explanation remains relevant. Hence, it is crucial to obtain an accurate and robust classifier with good generalization abilities to ensure reliable reliance on its explanations. 

To quantitatively evaluate and compare the growth monitoring abilities of the different XAI methods, we first assess whether the estimated severity metrics exhibit linear variations. This is done by computing the average $r$-value, representing the correlation coefficient of a linear fit of the estimated severity as a function of time. Secondly, we analyze whether the slopes of the estimated severity are in close agreement with the slopes of the ground-truth severity metric. This is assessed by calculating the mean absolute percentage error (MAPE) between the slopes. We filter the dataset to include only images classified as cracks by the classifier, and discard trajectories with fewer than three elements. This filtering process results in 87 retained growth trajectories out of the initial 100. The results for the area and maximum width metrics are reported in Table~\ref{tab:growth-results}.

\begin{table}
    \setlength{\aboverulesep}{0pt}
    \setlength{\belowrulesep}{0pt}
    \centering
    \caption{Assessment of crack growth monitoring abilities of different XAI methods in our proposed methodology, using 100 artificially generated linear growth trajectories. Severity metrics are the crack area and maximum width. The table reports the average $r$-value of a linear fit of the estimated severity as a function of time, and the mean absolute percentage error (MAPE, in \%) between the estimated slope (i.e., growth rate) and the ground-truth slope. Best and second-best scores in bold underlined and bold, respectively.}
    \label{tab:growth-results}
    {\small
    \begin{tabular}{|l|cc|cc|}
        \toprule
        \multirow{2}*{Method} & \multicolumn{2}{c|}{Area growth} & \multicolumn{2}{c|}{Width growth} \\
        & Average $r$ & Slope MAPE & Average $r$ & Slope MAPE \\
        \midrule
        Input$\times$Gradient & -0.32 & 235.9 & -0.03 & 110.1 \\
        IntGrad & 0.77 & 151.3 & 0.22 & \phantom{0}77.7 \\
        DeepLift & 0.44 & \phantom{0}\textbf{84.7} & 0.18 & \phantom{0}88.4 \\
        DeepLiftShap & \underline{\textbf{0.90}} & \phantom{0}88.0 & \textbf{0.71} & \phantom{0}\textbf{73.8} \\
        GradientShap & 0.33 & 367.7 & 0.07 & 109.1 \\
        LRP & \textbf{0.84} & \phantom{0}\underline{\textbf{35.6}} & \underline{\textbf{0.80}} & \phantom{0}\underline{\textbf{37.8}} \\
        NN-Explainer & -0.81 & 182.8 & -0.48 & 120.3 \\
        B-cos network & 0.71 & 259.7 & 0.28 & \phantom{0}92.4 \\
        \midrule
        \rev{U-Net (Oracle)} & \rev{0.99} & \rev{\phantom{0}11.7} & \rev{0.88} & \rev{\phantom{0}11.0} \\
        \bottomrule
    \end{tabular}
    }
\end{table}

We observe that only DeepLiftShap and LRP consistently exhibit a strong linear correlation for both the area and width metrics, with high $r$-values. IntGrad and the B-cos network achieve a high $r$-value for the area metric only. For the other methods, there was no clear positive correlation between crack growth and the metrics extracted from the XAI-based segmentation masks, indicating that the growth of the crack did not translate into the resulting attribution maps. However, we also observed a general degradation in the quality of the explanations for artificially generated cracks because their appearance does not closely resemble the real data distribution, which is a limitation of our study. In terms of slope, representing the severity growth rate, LRP achieves the best accuracy with an approximate 35\% MAPE. \rev{The supervised oracle achieves the highest $r$-values and the lowest error on the growth slopes, with 11.0 and 11.7 MAPE on the area and width, respectively.}

\begin{figure}
    \centering
    \includegraphics[width=0.6\textwidth]{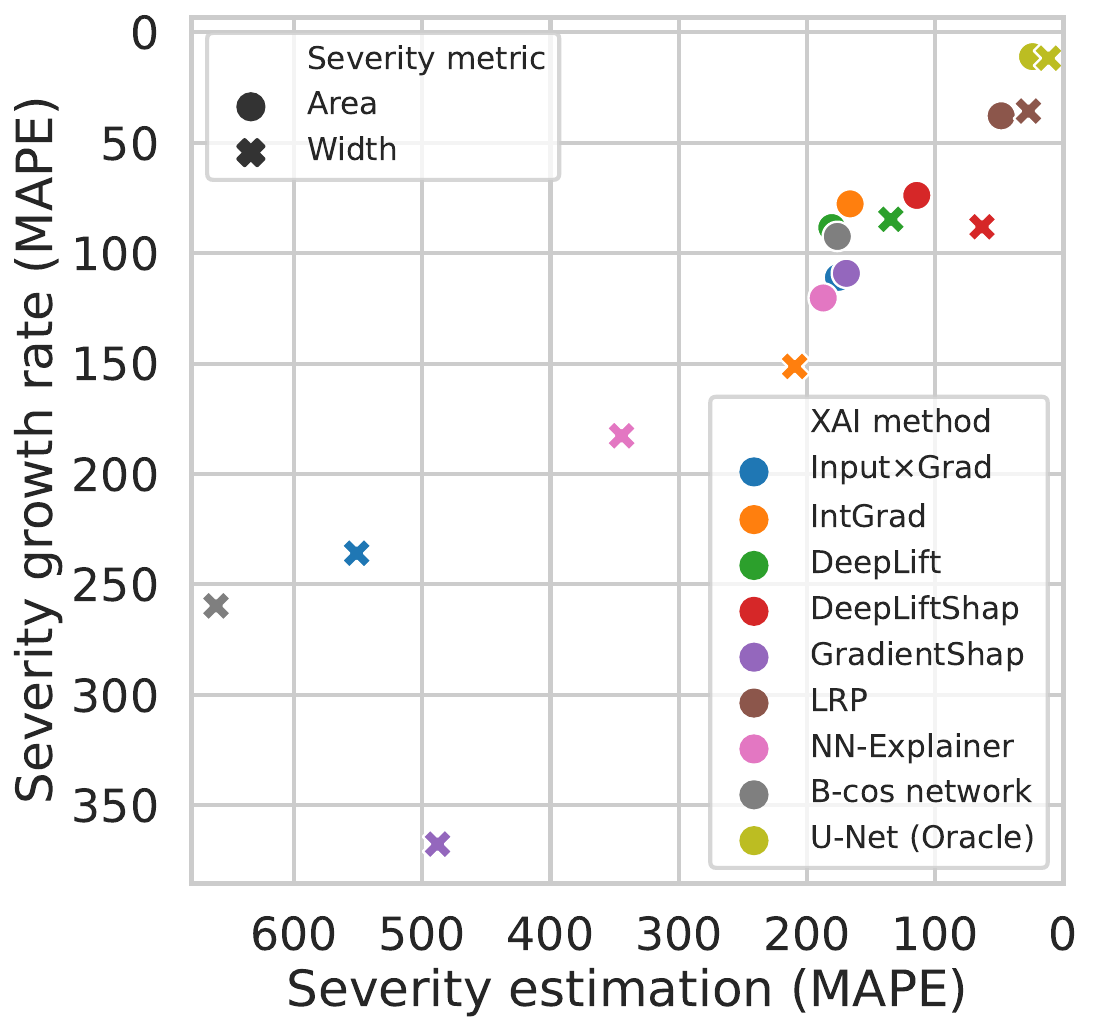}
    \caption{Comparison of the XAI methods used in our study in terms of crack severity quantification performance and crack severity growth rate estimation performance, using area and maximum width as severity metrics. Mean absolute percentage error compared with the ground-truth severity. Best methods lie in the upper right corner. \rev{LRP is closest to the oracle method (supervised U-Net).}}
    \label{fig:severity-plot}
\end{figure}

Finally, we summarize this experiment with a two-dimensional plot (Figure~\ref{fig:severity-plot}) representing the error in severity estimation on the x-axis and the error in severity growth rate estimation (i.e., linear slope) on the y-axis, for each of the XAI methods. A good method should be located in the upper right corner, indicating low errors in both dimensions. Among all the XAI methods compared in our study, LRP demonstrated the most consistent behavior in terms of severity quantification and growth monitoring. The second-best method is DeepLiftShap.

\subsection{Runtime comparison}

\begin{figure}
    
    \includegraphics[width=0.75\textwidth]{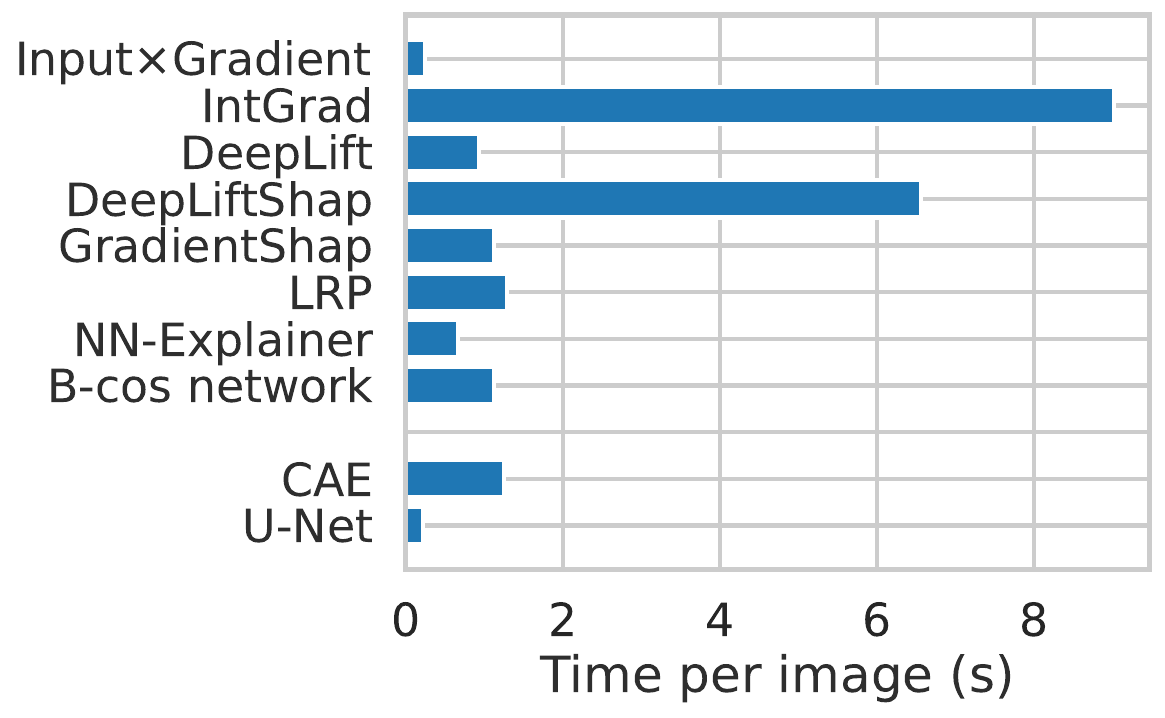}
    \caption{Comparison of computational runtimes on the DIC images (size $256 \times 256$).}
    \label{fig:runtimes}
\end{figure}

In this section, we compare the computational runtime performance of the evaluated  methods. Experiments were performed on a server with an Intel Xeon E5-2620 CPU, an NVIDIA GeForce RTX 2080 Ti GPU and running Ubuntu 22.04. The processing times per image in seconds are reported in Figure~\ref{fig:runtimes}.

Among the XAI methods, the fastest is Input$\times$Gradient at 0.251 s/image. The NN-Explainer requires a single forward pass of the \textit{explainer} network, making it the second-fastest method at 0.664 s/image. DeepLift, GradientShap, B-cos networks and LRP each take around one second per image. DeepLiftShap requires applying DeepLift for each sample of the baseline distribution, making it naturally slower and scaling linearly with the size of the baseline distribution (here with 10 samples). Integrated Gradients is the slowest method in our study, as it requires multiple steps to interpolate between the baseline and the input. We kept the default number of steps in the \texttt{captum} library, equal to 50. The post-processing time is negligible when using the simple thresholding strategy (0.018 s/image, including the morphological operations), but increases by one order of magnitude with the GMM thresholding (0.372 s/image).

In conclusion, the LRP method provides the best compromise in terms of segmentation quality, growth monitoring ability and computational runtime. In cases where LRP cannot be used, for instance, with classifier architectures containing skip-connections, DeepLiftShap provides an effective solution. Overall, the entire approach is slower but still in a similar order of magnitude as the inference of a supervised segmentation model such as U-Net.

\section{Conclusion}\label{sec:conclusion}

Automated segmentation and severity quantification of cracks in images are crucial tasks in structural health monitoring. However, deep learning-based semantic segmentation algorithms require extensive pixel-level labeling of large datasets for supervision. In this work, we have proposed a methodology and benchmarked the performance of various explainable artificial intelligence (XAI) methods, as well as post-processing techniques, in generating high-quality segmentation masks for cracks in masonry building wall surfaces. These masks are derived from the explanations of a binary classifier, trained on damage-free and damaged samples. Moreover, we have proposed a modification of the Neural Network Explainer method suitable for damage classification applications, where a negative class represents the absence of damage. Additionally, we have proposed using damage-free images as baselines in the Integrated Gradients and DeepLift-based XAI methods to enhance the quality of their explanations. The results of our benchmark study have demonstrated  that this methodology allows us to approximate segmentation masks with promising performance (around 0.4 F1-score). While this performance falls below that of fully supervised segmentation approaches such as U-Net, it outperforms purely unsupervised approaches such as autoencoders. 

Finally, we have investigated the applicability of these methods in quantifying damage severity and monitoring its progression. We evaluated severity metrics such as the number of cracks, maximum crack width and crack area. The experimental results demonstrated that accurately estimating the severity metrics was challenging, primarily due to overestimation of the crack extent. However, we found that it was possible to effectively monitor the severity evolution over time.  One key takeaway from this study is the effectiveness of the Layer-wise Relevance Propagation (LRP) method, which excelled in terms of segmentation quality, growth monitoring ability, and computational efficiency.

It is worth mentioning that there are other variations of this methodology that can be explored. For instance, the resulting masks could serve as approximate, coarse labels for training a supervised segmentation model. If ground-truth pixel-level labels are available, these coarse labels could be fine-tuned or used in a semi-supervised setting. These strategies are left for future work. 

\rev{We believe this approach holds the potential to significantly accelerate the development of automated crack detection and monitoring systems by eliminating  the need for time-consuming and costly pixel-level image annotation. However, it is important to note that in scenarios where related labeled datasets are available for supervised segmentation,  transfer learning is likely to produce more accurate results. Therefore, the methodology presented in this work represents an alternative in scenarios where transfer learning is challenging, e.g., when dealing with various types of damages and material aspects for which no public supervised segmentation datasets are available.}

As part of future work, we also plan to apply the methodology to different types of defects and different types of infrastructure, such as railway sleepers. Moreover, we aim to evaluate the approach using real crack growth data. Finally, we intend to investigate other families of explainable AI methods, beyond feature attribution explanation methods.

\section*{Acknowledgment}
This work was supported by the EPFL ENAC Cluster grant "explAIn". We also thank the EPFL EESD lab for providing additional images.

\appendix
\color{black}

\section{Additional results with different architectures}

In this appendix, we report additional results for different classifier architectures, namely ResNet-18 and ViT-B/16. Note that the post-processing steps have not been tuned for these models; results are not optimal and serve as a demonstration that the methodology is applicable for various classifiers beyond the VGG used in the study.

\subsection{Additional results with ResNet-18 classifier}

\begin{table}[H]
    \centering
    \caption{\rev{Crack classification performance on the DIC dataset (values in \%).}}\label{tab:classifier-results-resnet}
    {\small
    \begin{tabular}{>{\color{black}}l>{\color{black}}l>{\color{black}}c>{\color{black}}c>{\color{black}}c}
    \toprule
     Model & Fold & Balanced accuracy & TPR & TNR \\
     \midrule
     \multirow{3}*{ResNet-18} & train & 100.0 & 100.0 & 100.0 \\
     & val & \phantom{0}92.0 & \phantom{0}86.0 & \phantom{0}98.0 \\
     & test & \phantom{0}86.5 & \phantom{0}75.0 & \phantom{0}98.1 \\
     \bottomrule
    \end{tabular}
    }
\end{table}

\begin{table}[H]
    \setlength{\aboverulesep}{0pt}
    \setlength{\belowrulesep}{0pt}
    \centering
    \caption{\rev{Crack segmentation quality obtained with the proposed weakly-supervised methodology using different explainable AI methods and post-processing techniques (values in \%). Best and second-best scores in bold underlined and bold, respectively.}}
    \label{tab:segmentation-results-resnet}
    {\scriptsize
    \begin{tabular}{|>{\color{black}}l|>{\color{black}}l|>{\color{black}}c>{\color{black}}c|>{\color{black}}c>{\color{black}}c>{\color{black}}c>{\color{black}}c|}
        \toprule
        \multicolumn{2}{|c|}{\multirow{2}*{\rev{Method}}} & \multicolumn{2}{c|}{\rev{Post-processing}} & \multirow{2}*{F1} & \multirow{2}*{Precision} & \multirow{2}*{Recall} & \multirow{2}*{IoU} \\
        \multicolumn{2}{|c|}{} & \rev{thresh.} & \rev{morph.} & & & & \\
        \midrule
        \multirow{24}*{\rotatebox[origin=c]{90}{\rev{XAI-based (weakly-supervised) -- ResNet18}}}
        & \multirow{4}*{\rev{Input$\times$Gradient}} & \rev{simple} & \rev{\xmark} & \rev{\phantom{0}5.01} & \rev{\phantom{0}2.79} & \rev{24.53} & \rev{\phantom{0}2.57} \\
         &  & \rev{simple} & \rev{\cmark} & \rev{\phantom{0}5.87} & \rev{\phantom{0}3.03} & \rev{97.16} & \rev{\phantom{0}3.03} \\
         &  & \rev{GMM} & \rev{\xmark} & \rev{\phantom{0}4.19} & \rev{\phantom{0}2.93} & \rev{\phantom{0}7.33} & \rev{\phantom{0}2.14} \\
         &  & \rev{GMM} & \rev{\cmark} & \rev{\phantom{0}8.43} & \rev{\phantom{0}4.77} & \rev{35.99} & \rev{\phantom{0}4.40} \\
        \cmidrule{2-8}
         & \multirow{4}*{\rev{IntGrad}} & \rev{simple} & \rev{\xmark} & \rev{\phantom{0}9.77} & \rev{\phantom{0}5.66} & \rev{35.62} & \rev{\phantom{0}5.13} \\
         & & \rev{simple} & \rev{\cmark} & \rev{\phantom{0}7.11} & \rev{\phantom{0}3.69} & \rev{\underline{\textbf{99.09}}} & \rev{\phantom{0}3.68} \\
         & & \rev{GMM} & \rev{\xmark} & \rev{17.35} & \rev{12.86} & \rev{26.67} & \rev{\phantom{0}9.50} \\
         & & \rev{GMM} & \rev{\cmark} & \rev{22.86} & \rev{13.68} & \rev{69.64} & \rev{12.91} \\
        \cmidrule{2-8}
         & \multirow{4}*{\rev{DeepLift}} & \rev{simple} & \rev{\xmark} & \rev{15.41} & \rev{10.21} & \rev{31.42} & \rev{\phantom{0}8.35} \\
         & & \rev{simple} & \rev{\cmark} & \rev{11.97} & \rev{\phantom{0}6.42} & \rev{88.64} & \rev{\phantom{0}6.37} \\
         & & \rev{GMM} & \rev{\xmark} & \rev{26.89} & \rev{\textbf{26.99}} & \rev{26.79} & \rev{15.53} \\
         & & \rev{GMM} & \rev{\cmark} & \rev{\textbf{36.08}} & \rev{26.41} & \rev{56.90} & \rev{\textbf{22.01}} \\
        \cmidrule{2-8}
         & \multirow{4}*{\rev{DeepLiftshap}} & \rev{simple} & \rev{\xmark} & \rev{\phantom{0}9.92} & \rev{\phantom{0}5.66} & \rev{40.32} & \rev{\phantom{0}5.22} \\
         & & \rev{simple} & \rev{\cmark} & \rev{\phantom{0}7.04} & \rev{\phantom{0}3.65} & \rev{98.39} & \rev{\phantom{0}3.65} \\
         & & \rev{GMM} & \rev{\xmark} & \rev{26.47} & \rev{25.83} & \rev{27.14} & \rev{15.25} \\
         & & \rev{GMM} & \rev{\cmark} & \rev{\underline{\textbf{37.54}}} & \rev{\underline{\textbf{27.04}}} & \rev{61.34} & \rev{\underline{\textbf{23.11}}} \\
        \cmidrule{2-8}
         & \multirow{4}*{\rev{GradientShap}} & \rev{simple} & \rev{\xmark} & \rev{\phantom{0}5.57} & \rev{\phantom{0}2.96} & \rev{48.23} & \rev{\phantom{0}2.87} \\
         & & \rev{simple} & \rev{\cmark} & \rev{\phantom{0}5.81} & \rev{\phantom{0}2.99} & \rev{\textbf{98.81}} & \rev{\phantom{0}2.99} \\
         & & \rev{GMM} & \rev{\xmark} & \rev{11.63} & \rev{\phantom{0}8.65} & \rev{17.77} & \rev{\phantom{0}6.17} \\
         & & \rev{GMM} & \rev{\cmark} & \rev{18.70} & \rev{11.52} & \rev{49.61} & \rev{10.31} \\
        \cmidrule{2-8}
         & \multirow{4}*{\rev{LRP}} & \rev{simple} & \rev{\xmark} & \rev{26.23} & \rev{16.76} & \rev{60.24} & \rev{15.09} \\
         & & \rev{simple} & \rev{\cmark} & \rev{23.89} & \rev{14.23} & \rev{74.54} & \rev{13.57} \\
         & & \rev{GMM} & \rev{\xmark} & \rev{28.31} & \rev{20.42} & \rev{46.12} & \rev{16.49} \\
         & & \rev{GMM} & \rev{\cmark} & \rev{27.70} & \rev{18.50} & \rev{55.11} & \rev{16.08} \\
        \midrule
        \multirow{8}*{\rotatebox[origin=c]{90}{Unsupervised}}
        & \multirow{4}*{Raw} & simple & \xmark & 12.18 & \phantom{0}6.54 & 89.46 & \phantom{0}6.49 \\
        & & simple & \cmark & \phantom{0}4.73 & \phantom{0}2.42 & 100.0 & \phantom{0}2.42 \\
        & & GMM & \xmark & 18.05 & 10.21 & 78.00 & \phantom{0}9.92 \\
        & & GMM & \cmark & \phantom{0}7.88 & \phantom{0}4.12 & 91.36 & \phantom{0}4.10 \\
        \cmidrule{2-8}
        & \multirow{4}*{CAE} & \rev{simple} & \xmark & 12.39 & \phantom{0}7.05 & 51.22 & \phantom{0}6.60 \\
        & & \rev{simple} & \cmark & \phantom{0}5.93 & \phantom{0}3.07 & 90.09 & \phantom{0}3.06 \\
        & & \rev{GMM} & \xmark & 15.36 & \phantom{0}9.39 & 42.11 & \phantom{0}8.32 \\
        & & \rev{GMM} & \cmark & 13.32 & \phantom{0}7.57 & 55.68 & \phantom{0}7.14 \\
        \midrule
        \multicolumn{4}{|l|}{\rev{Supervised U-Net (oracle)}} & 83.67 & 82.22 & 85.17 & 71.93 \\
        \bottomrule
    \end{tabular}
    }
\end{table}

\begin{table}
    \setlength{\aboverulesep}{0pt}
    \setlength{\belowrulesep}{0pt}
    \centering
    \caption{\rev{Assessment of crack severity estimation using number of cracks per patch (CPP), crack area per patch and maximum crack width. The table reports the mean absolute error (MAE) or mean absolute percentage error (MAPE, in \%) with the ground-truth severity measure (lower is better). Best and second-best scores in bold underlined and bold, respectively.}}
    \label{tab:severity-results-resnet}
    {\scriptsize
    \begin{tabular}{|>{\color{black}}l|>{\color{black}}l|>{\color{black}}c>{\color{black}}c|>{\color{black}}c>{\color{black}}c>{\color{black}}c|}
        \toprule
        \multicolumn{2}{|c|}{\multirow{2}*{\rev{Method}}} & \multicolumn{2}{c|}{\rev{Post-processing}} & \rev{CPP} & Area & Width \\
        \multicolumn{2}{|c|}{} & \rev{thresh.} & \rev{morph.} & \rev{MAE} & MAPE & MAPE (MAE px) \\
        \midrule
        \multirow{12}*{\rotatebox[origin=c]{90}{\rev{XAI-based -- ResNet18}}}
        & \multirow{2}*{\rev{Input$\times$Gradient}} 
         & \rev{simple} & \cmark & \textbf{1.01} & 5959.0 & \textbf{201.0} (17.44) \\
         & & \rev{GMM} & \cmark & 3.07 & 1144.2 & 371.1 (28.29) \\
        \cmidrule{2-7}
        & \multirow{2}*{\rev{IntGrad}} 
         & \rev{simple} & \cmark & 1.41 & 5346.8 & 258.8 (21.39) \\
         & & \rev{GMM} & \cmark & 2.25 & 916.0 & 361.2 (26.17) \\
        \cmidrule{2-7}
        & \multirow{2}*{\rev{DeepLift}} 
         & \rev{simple} & \cmark & 1.60 & 2409.3 & 315.9 (25.27) \\
         & & \rev{GMM} & \cmark & 1.40 & \textbf{367.4} & 299.2 (21.40) \\
        \cmidrule{2-7}
        & \multirow{2}*{\rev{DeepLiftShap}} 
         & \rev{simple} & \cmark & 1.33 & 5032.2 & 219.2 (19.27) \\
         & & \rev{GMM} & \cmark & 1.43 & \underline{\textbf{350.7}} & 299.6 (21.56) \\
        \cmidrule{2-7}
        & \multirow{2}*{\rev{GradientShap}} 
         & \rev{simple} & \cmark & 1.05 & 6263.1 & \underline{\textbf{113.4}} (12.01) \\
         & & \rev{GMM} & \cmark & 3.25 & 751.1 & 379.7 (28.28) \\
        \cmidrule{2-7}
        & \multirow{2}*{\rev{LRP}} 
         & \rev{simple} & \cmark & \underline{\textbf{0.84}} & 906.3 & 357.6 (27.37) \\
         & & \rev{GMM} & \cmark & 1.21 & 488.3 & 348.7 (26.29) \\
        \midrule
        \multicolumn{4}{|l|}{\rev{Supervised U-Net (oracle)}} & 0.74 & \phantom{0}20.1 & \phantom{0}20.8 (1.67) \\
        \bottomrule
    \end{tabular}
    }
\end{table}

\clearpage
\subsection{Additional results with ViT-B/16 classifier}

\begin{table}[H]
    \centering
    \caption{\rev{Crack classification performance on the DIC dataset (values in \%).}}\label{tab:classifier-results-vit}
    {\small
    \begin{tabular}{>{\color{black}}l>{\color{black}}l>{\color{black}}c>{\color{black}}c>{\color{black}}c}
    \toprule
     Model & Fold & Balanced accuracy & TPR & TNR \\
     \midrule
     \multirow{3}*{ViT-B/16} & train & \phantom{0}99.3 & \phantom{0}98.7 & 100.0 \\
     & val & \phantom{0}90.7 & \phantom{0}84.5 & \phantom{0}97.0 \\
     & test & \phantom{0}85.8 & \phantom{0}74.0 & \phantom{0}97.6 \\
     \bottomrule
    \end{tabular}
    }
\end{table}

\begin{table}[H]
    \setlength{\aboverulesep}{0pt}
    \setlength{\belowrulesep}{0pt}
    \centering
    \caption{\rev{Crack segmentation quality obtained with the proposed weakly-supervised methodology using different explainable AI methods and post-processing techniques (values in \%). Best and second-best scores in bold underlined and bold, respectively.}}
    \label{tab:segmentation-results-vit}
    {\scriptsize
    \begin{tabular}{|>{\color{black}}l|>{\color{black}}l|>{\color{black}}c>{\color{black}}c|>{\color{black}}c>{\color{black}}c>{\color{black}}c>{\color{black}}c|}
        \toprule
        \multicolumn{2}{|c|}{\multirow{2}*{\rev{Method}}} & \multicolumn{2}{c|}{\rev{Post-processing}} & \multirow{2}*{F1} & \multirow{2}*{Precision} & \multirow{2}*{Recall} & \multirow{2}*{IoU} \\
        \multicolumn{2}{|c|}{} & \rev{thresh.} & \rev{morph.} & & & & \\
        \midrule
        \multirow{24}*{\rotatebox[origin=c]{90}{\rev{XAI-based (weakly-supervised) -- ViT-B/16}}}
        & \multirow{4}*{\rev{Input$\times$Gradient}} & \rev{simple} & \rev{\xmark} & \rev{\phantom{0}8.60} & \rev{\phantom{0}6.15} & \rev{14.30} & \rev{\phantom{0}4.49} \\
        & & \rev{simple} & \rev{\cmark} & \rev{13.16} & \rev{\phantom{0}7.25} & \rev{\underline{\textbf{71.06}}} & \rev{\phantom{0}7.04} \\
        & & \rev{GMM} & \rev{\xmark} & \rev{10.27} & \rev{10.75} & \rev{\phantom{0}9.83} & \rev{\phantom{0}5.41} \\
        & & \rev{GMM} & \rev{\cmark} & \rev{21.83} & \rev{16.33} & \rev{32.91} & \rev{12.25} \\
        \cmidrule{2-8}
        & \multirow{4}*{\rev{IntGrad}} & \rev{simple} & \rev{\xmark} & \rev{25.24} & \rev{37.04} & \rev{19.15} & \rev{14.45} \\
        & & \rev{simple} & \rev{\cmark} & \rev{\textbf{40.81}} & \rev{32.52} & \rev{54.75} & \rev{\textbf{25.63}} \\
        & & \rev{GMM} & \rev{\xmark} & \rev{38.73} & \rev{\underline{\textbf{42.91}}} & \rev{35.30} & \rev{24.02} \\
        & & \rev{GMM} & \rev{\cmark} & \rev{\underline{\textbf{46.83}}} & \rev{\textbf{37.16}} & \rev{63.28} & \rev{\underline{\textbf{30.57}}} \\
        \cmidrule{2-8}
        & \multirow{4}*{\rev{DeepLift}} & \rev{simple} & \rev{\xmark} & \rev{16.12} & \rev{15.70} & \rev{16.57} & \rev{\phantom{0}8.77} \\
        & & \rev{simple} & \rev{\cmark} & \rev{22.14} & \rev{13.78} & \rev{56.34} & \rev{12.45} \\
        & & \rev{GMM} & \rev{\xmark} & \rev{30.07} & \rev{30.72} & \rev{29.44} & \rev{17.69} \\
        & & \rev{GMM} & \rev{\cmark} & \rev{38.75} & \rev{29.15} & \rev{57.79} & \rev{24.03} \\
        \cmidrule{2-8}
        & \multirow{4}*{\rev{DeepLiftshap}} & \rev{simple} & \rev{\xmark} & \rev{14.97} & \rev{14.99} & \rev{14.94} & \rev{\phantom{0}8.09} \\
        & & \rev{simple} & \rev{\cmark} & \rev{22.45} & \rev{14.26} & \rev{52.74} & \rev{12.64} \\
        & & \rev{GMM} & \rev{\xmark} & \rev{27.23} & \rev{28.64} & \rev{25.95} & \rev{15.76} \\
        & & \rev{GMM} & \rev{\cmark} & \rev{37.63} & \rev{28.91} & \rev{53.86} & \rev{23.17} \\
        \cmidrule{2-8}
        & \multirow{4}*{\rev{GradientShap}} & \rev{simple} & \rev{\xmark} & \rev{15.71} & \rev{17.79} & \rev{14.07} & \rev{\phantom{0}8.52} \\
        & & \rev{simple} & \rev{\cmark} & \rev{24.20} & \rev{16.14} & \rev{48.37} & \rev{13.77} \\
        & & \rev{GMM} & \rev{\xmark} & \rev{21.11} & \rev{32.35} & \rev{15.66} & \rev{11.80} \\
        & & \rev{GMM} & \rev{\cmark} & \rev{32.35} & \rev{32.77} & \rev{31.94} & \rev{19.29} \\
        \cmidrule{2-8}
        & \multirow{4}*{\rev{LRP-ViT*}} & \rev{simple} & \rev{\xmark} & \rev{18.71} & \rev{11.13} & \rev{58.54} & \rev{10.32} \\
         & & \rev{simple} & \rev{\cmark} & \rev{17.29} & \rev{\phantom{0}9.97} & \rev{\textbf{65.03}} & \rev{\phantom{0}9.46} \\
         & & \rev{GMM} & \rev{\xmark} & \rev{24.08} & \rev{16.30} & \rev{46.03} & \rev{13.69} \\
         & & \rev{GMM} & \rev{\cmark} & \rev{23.50} & \rev{15.30} & \rev{50.68} & \rev{13.31} \\
        \midrule
        \multirow{8}*{\rotatebox[origin=c]{90}{Unsupervised}}
        & \multirow{4}*{Raw} & simple & \xmark & 12.18 & \phantom{0}6.54 & 89.46 & \phantom{0}6.49 \\
        & & simple & \cmark & \phantom{0}4.73 & \phantom{0}2.42 & 100.0 & \phantom{0}2.42 \\
        & & \rev{GMM} & \xmark & 18.05 & 10.21 & 78.00 & \phantom{0}9.92 \\
        & & \rev{GMM} & \cmark & \phantom{0}7.88 & \phantom{0}4.12 & 91.36 & \phantom{0}4.10 \\
        \cmidrule{2-8}
        & \multirow{4}*{CAE} & simple & \xmark & 12.39 & \phantom{0}7.05 & 51.22 & \phantom{0}6.60 \\
        & & simple & \cmark & \phantom{0}5.93 & \phantom{0}3.07 & 90.09 & \phantom{0}3.06 \\
        & & \rev{GMM} & \xmark & 15.36 & \phantom{0}9.39 & 42.11 & \phantom{0}8.32 \\
        & & \rev{GMM} & \cmark & 13.32 & \phantom{0}7.57 & 55.68 & \phantom{0}7.14 \\
        \midrule
        \multicolumn{4}{|l|}{\rev{Supervised U-Net (oracle)}} & 83.67 & 82.22 & 85.17 & 71.93 \\
        \bottomrule
    \end{tabular}
    }
\end{table}

\begin{table}
    \setlength{\aboverulesep}{0pt}
    \setlength{\belowrulesep}{0pt}
    \centering
    \caption{\rev{Assessment of crack severity estimation using number of cracks per patch (CPP), crack area per patch and maximum crack width. The table reports the mean absolute error (MAE) or mean absolute percentage error (MAPE, in \%) with the ground-truth severity measure (lower is better). Best and second-best scores in bold underlined and bold, respectively.}}
    \label{tab:severity-results-vit}
    {\scriptsize
    \begin{tabular}{|>{\color{black}}l|>{\color{black}}l|>{\color{black}}c>{\color{black}}c|>{\color{black}}c>{\color{black}}c>{\color{black}}c|}
        \toprule
        \multicolumn{2}{|c|}{\multirow{2}*{\rev{Method}}} & \multicolumn{2}{c|}{\rev{Post-processing}} & \rev{CPP} & Area & Width \\
        \multicolumn{2}{|c|}{} & \rev{thresh.} & \rev{morph.} & \rev{MAE} & \rev{MAPE} & \rev{MAPE (MAE px)} \\
        \midrule
        \multirow{12}*{\rotatebox[origin=c]{90}{\rev{XAI-based -- ViT-B/16}}}
        & \multirow{2}*{\rev{Input$\times$Gradient}} 
        & simple & \cmark & 1.86 & 1134.4 & 270.7 (21.89) \\
        & & \rev{GMM} & \cmark & 2.62 & 276.5 & 289.8 (20.78) \\
        \cmidrule{2-7}
        & \multirow{2}*{\rev{IntGrad}} 
        & \rev{simple} & \cmark & \textbf{1.00} & \textbf{151.9} & \underline{\textbf{194.6}} (14.26) \\
        & & \rev{GMM} & \cmark & \underline{\textbf{0.84}} & 201.9 & 268.5 (18.76) \\
        \cmidrule{2-7}
        & \multirow{2}*{\rev{DeepLift}} 
        & \rev{simple} & \cmark & 2.26 & 529.0 & 252.3 (18.97) \\
        & & \rev{GMM} & \cmark & 1.89 & 265.7 & 284.6 (19.82) \\
        \cmidrule{2-7}
        & \multirow{2}*{\rev{DeepLiftShap}} 
        & \rev{simple} & \cmark & 2.47 & 462.2 & 254.6 (18.86) \\
        & & \rev{GMM} & \cmark & 2.07 & 249.6 & 294.0 (19.96) \\
        \cmidrule{2-7}
        & \multirow{2}*{\rev{GradientShap}} 
        & \rev{simple} & \cmark & 1.62 & 558.7 & \textbf{200.7} (14.95) \\
        & & \rev{GMM} & \cmark & 1.16 & \phantom{0}\underline{\textbf{93.5}} & 210.4 (15.03) \\
        \cmidrule{2-7}
        & \multirow{2}*{\rev{LRP-ViT*}} 
         & \rev{simple} & \cmark & 4.30 & 1351.8 & 394.2 (25.8) \\
         & & \rev{GMM} & \cmark & 3.35 & 579.9 & 319.8 (20.9) \\
        \midrule
        \multicolumn{4}{|l|}{\rev{Supervised U-Net (oracle)}} & \rev{0.74} & \rev{\phantom{0}20.1} & \rev{\phantom{0}20.8 (1.67)} \\
        \bottomrule
    \end{tabular}
    }
\end{table}

\rev{We use the LRP adaptation for ViT by \cite{chefer_transformer_2021}, which outputs attribution maps of size $s \times s$ where $s$ is the number of patch tokens. As we use ViT-B/16 with patch size 16px and image size 224px, the maps are of low resolution, i.e. $14\times 14$, and upscaled to the original image resolution, which explains poor performance metrics. In real applications, a model with smaller patches (e.g., 8px or smaller) should be used to obtain high-resolution maps, at an increased computational and memory cost.}

\color{black}


 \bibliographystyle{elsarticle-num} 
 \bibliography{references_clean}





\end{document}